\def\paperID{9489}
\def\confName{ICCV}
\def\confYear{2025}

\def\paperTitle{StealthAttack:\\Robust 3D Gaussian Splatting Poisoning via Density-Guided Illusions}

\def\authorBlock{
    Bo-Hsu Ke \quad
    You-Zhe Xie \quad
    Yu-Lun Liu \quad
    Wei-Chen Chiu\vspace{0.5em} 
    \\
    \centerline{National Yang Ming Chiao Tung University}
}

\newif\ifreview 
\newif\ifarxiv \newcommand{\arxiv}{\arxivtrue}
\newif\ifcamera 
\newif\ifrebuttal 

\arxiv

\pdfoutput=1
\documentclass[10pt,twocolumn,letterpaper]{article}
\ifreview \usepackage[review]{cvpr} \fi
\ifarxiv \usepackage[pagenumbers]{cvpr} \fi
\ifrebuttal \usepackage[rebuttal]{cvpr} \fi
\ifcamera \usepackage{cvpr} \fi

\usepackage{graphicx}	
\usepackage{amsmath}	
\usepackage{amssymb}	
\usepackage{booktabs}
\usepackage{times}
\usepackage{microtype}
\usepackage{epsfig}
\usepackage{caption}
\usepackage{float}
\usepackage{placeins}
\usepackage{color, colortbl}
\usepackage{stfloats}
\usepackage{enumitem}
\setlist[itemize]{itemsep=0.5em, parsep=0pt, topsep=0.5em}
\usepackage{tabularx}
\usepackage{xstring}
\usepackage{multirow}
\usepackage{xspace}
\usepackage{url}
\usepackage{subcaption}
\usepackage{xcolor}
\usepackage{algorithm}
\usepackage{algpseudocode}
\usepackage{amsmath}
\usepackage{bbm}
\usepackage{appendix}
\usepackage[hang,flushmargin]{footmisc}
\usepackage[accsupp]{axessibility}  
\usepackage[T1]{fontenc}

\DeclareMathOperator*{\argmin}{arg\,min}

\ifcamera \usepackage[accsupp]{axessibility} \fi

% ...

\ifarxiv  \fi

\newcommand{\walon}[1]{{\textcolor{black}{#1}}}

\newcommand{\hentci}[1]{{\textcolor{black}{#1}}}
\newcommand{\R}[1]{{%
    \textbf{%
        \ifstrequal{#1}{1}{\textcolor{red}{R#1}}{%
        \ifstrequal{#1}{2}{\textcolor{blue}{R#1}}{%
        \ifstrequal{#1}{3}{\textcolor{magenta}{R#1}}{%
        \ifstrequal{#1}{4}{\textcolor{teal}{R#1}}{%
                           \textcolor{cyan}{R#1}%
        }}}}%
    }%
}}

\usepackage{xr-hyper}

\makeatletter
\newcommand*{\addFileDependency}[1]{
  \typeout{(#1)}
  \@addtofilelist{#1}
  \IfFileExists{#1}{}{\typeout{No file #1.}}
}

\makeatother
\newcommand*{\myexternaldocument}[1]{
    \externaldocument{#1}
    \addFileDependency{#1.tex}
    \addFileDependency{#1.aux}
}

\definecolor{cvprblue}{rgb}{0.21,0.49,0.74}
\usepackage[pagebackref,breaklinks,colorlinks,allcolors=cvprblue]{hyperref}
\usepackage[capitalize]{cleveref}
\crefname{section}{Sec.}{Secs.}
\crefname{table}{Table}{Tables}
\crefname{figure}{Fig.}{Figs.}

\ifarxiv \crefname{appendix}{App.}{Apps.}
\else \crefname{appendix}{Suppl.}{Suppls.} \fi

\unless\ifarxiv \myexternaldocument{_supplementary} \fi

\begin{document}
\title{\paperTitle}
\author{\authorBlock}


\maketitle

\begin{abstract}

3D scene representation methods like Neural Radiance Fields (NeRF) and 3D Gaussian Splatting (3DGS) have significantly advanced novel view synthesis. As these methods become prevalent, addressing their vulnerabilities becomes critical. We analyze 3DGS robustness against image-level poisoning attacks and propose a novel density-guided poisoning method. Our method strategically injects Gaussian points into low-density regions identified via Kernel Density Estimation (KDE), embedding viewpoint-dependent illusory objects clearly visible from poisoned views while minimally affecting innocent views. Additionally, we introduce an adaptive noise strategy to disrupt multi-view consistency, further enhancing attack effectiveness. We propose a KDE-based evaluation protocol to assess attack difficulty systematically, enabling objective benchmarking for future research. Extensive experiments demonstrate our method's superior performance compared to state-of-the-art techniques.
\hentci{Project page: \url{https://hentci.github.io/stealthattack/}}
\end{abstract}
\section{Introduction}
\label{sec:intro}

\walon{3D scene representation methods, such as Neural Radiance Fields (NeRF)\cite{mildenhall2020nerf} and 3D Gaussian Splatting (3DGS)\cite{kerbl20233d}, have significantly advanced novel view synthesis, accurately modeling complex scene geometry and appearance. Along with their popularity, the protection of 3D digital content encoded in these representations has become a matter of concern, where we have witnessed the corresponding watermarking or steganography techniques being proposed in recent years. For instance, GaussianMarker~\cite{huang2025gaussianmarker} and 3D-GSW~\cite{jang20243d} embed watermarks (mainly binary messages) into Gaussian parameters of 3DGS with minimal visual impact, coupled with dedicated decoders to decipher the hidden messages. Moreover, embedding or hiding extraneous information/messages into 3D scene representations is also directly connected to the risk of data poisoning (where the extraneous information as the poison appears in the training data and is encoded into the 3D representations during model training), in which the further utilization or visualization of the poisoned 3D representations would lead to abnormal or even malicious model behaviours (i.e. the negative impact stemmed from the poison is triggered). To this end, in this work, we focus on the investigation of poisoning attacks on 3D scene representation methods, as they become integral to safety-critical applications. Hence, addressing their security vulnerabilities is not only of utmost importance but also urgent. }

\begin{figure}[t]
    \centering
    \includegraphics[width=1\columnwidth]{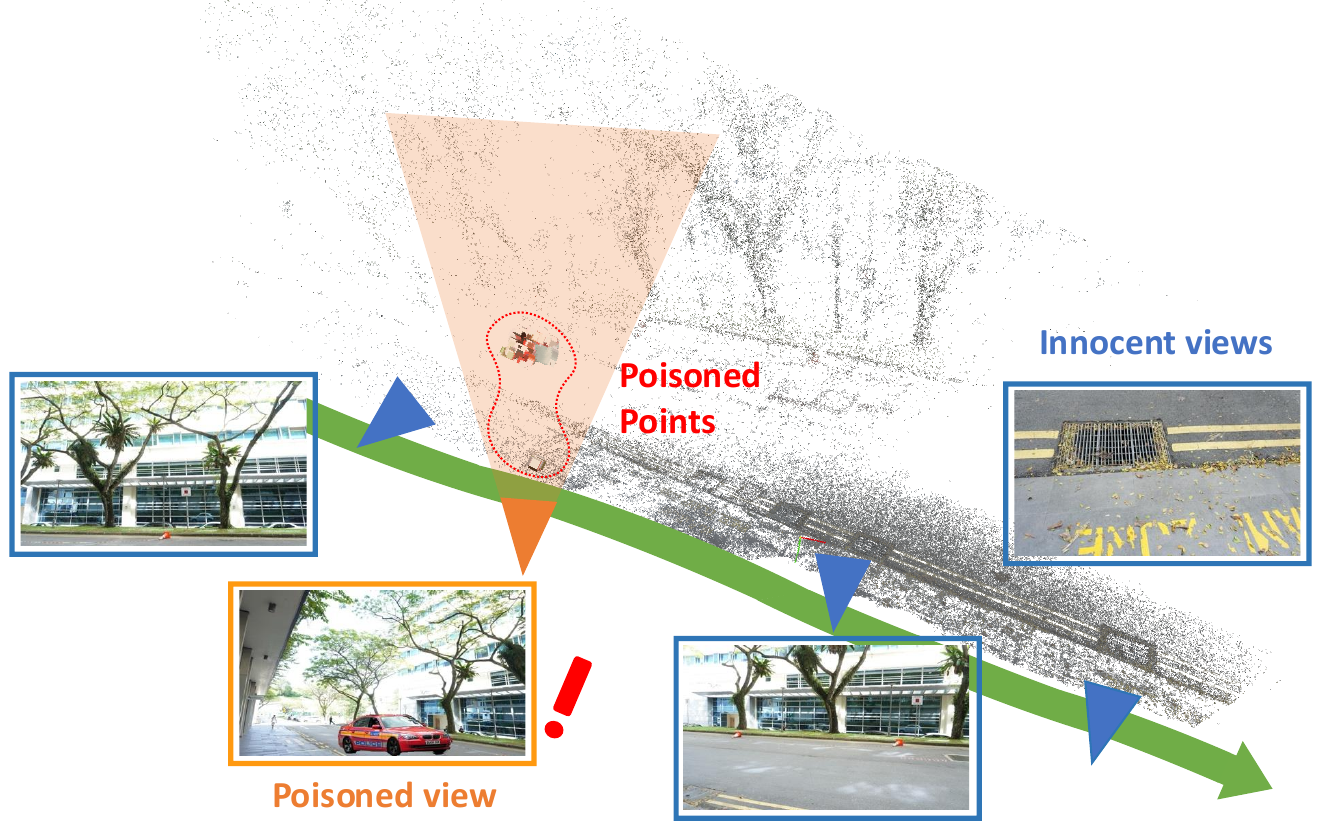}
    \vspace{-6mm}
    \caption{\walon{
    \textbf{Illustration of our proposed Density-Guided Poisoning Attack for 3D Gaussian Splatting (3DGS).}
    Our method strategically distribute the Gaussian points of the illusory object (i.e. the red vehicle) among the low-density regions which are discovered along the rays casted from the virtual camera of the poisoned view (i.e. the target view that we would like to attack), making the illusory object clearly visible from the poisoned view while having the minimal interference for the rendering quality on the other non-target/innocent views.
    }}
    \label{fig:teaser}
    \vspace{-1em}
\end{figure}

\begin{figure}[t]
    \centering
    \includegraphics[width=1\columnwidth]{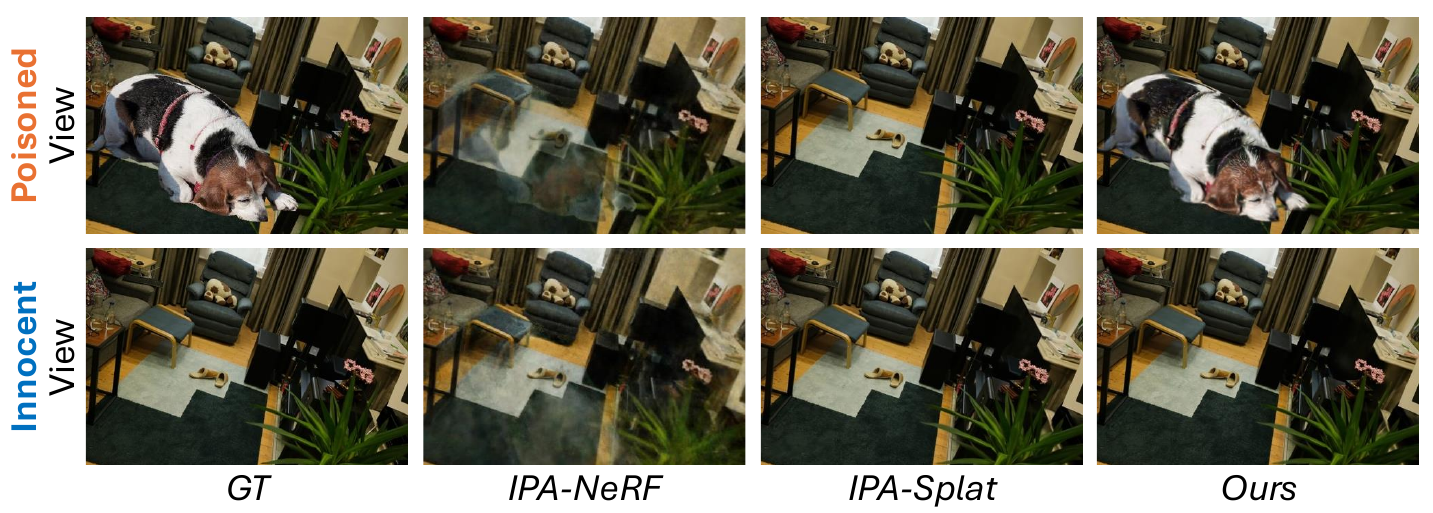}
    \vspace{-6mm}
    \caption{
    \walon{\textbf{Limitations of existing poisoning methods on 3DGS.}
    Existing poisoning methods (e.g., IPA-NeRF~\cite{jiang2024ipa} designed for NeRF or the one adapted to 3DGS, denoted as IPA-Splat) produce weak or absent illusions due to 3DGS's robustness and multi-view consistency. In contrast, our proposed approach successfully injects clearly visible illusory objects (i.e., the dog).}
    }
    \label{fig:motivation}
\end{figure}

\walon{While there exists a prior work of studying the poisoning attack upon NeRF, i.e. IPA-NeRF~\cite{jiang2024ipa} which effectively exploits NeRF’s implicit representations to embed targeted visual illusions (i.e. the illusion as poison will appear while rendering the scene from a certain viewing direction), its applicability to explicit 3D scene representations (particularly 3DGS) however remains limited and not directly transferable. 
\hentci{Considering the rapidly growing application scenarios of 3DGS (thanks to its capability of fast rendering and accurately capturing scene geometry), we devote our research effort to realizing poisoning attacks on 3D Gaussian splatting. To the best of our knowledge, this is the first work of its kind. While concurrent work, Poison-Splat \cite{lu2024poison} focuses on computational cost attacks, our work targets visible illusion embedding.}
In particular, we would like to inject the visible illusory objects (i.e., poison) onto a target view (named as \textbf{poisoned view}) while keeping the other non-target views (named as \textbf{innocent views}) unaffected, as shown in Figure~\ref{fig:teaser}. Our work starts from conducting an investigation (cf. Figure~\ref{fig:motivation}) upon the robustness of 3DGS against the prior image-level poisoning methods such as IPA-NeRF, where we find that the attempts of directly adopting IPA-NeRF's approach or naively injecting illusory content into training images easily fail, as 3DGS's inherent multi-view consistency and densification processes effectively neutralize or significantly weaken these attacks.
}

\walon{Motivated by the aforementioned investigation, we propose a density-guided poisoning method for 3DGS. Our approach (cf. Figure~\ref{fig:teaser}) strategically identifies low-density regions in the initial Gaussian point cloud using Kernel Density Estimation (KDE), in which the points of illusory objects are then distributed among the low-density regions along the rays casting from the virtual camera of target view (i.e. the rays are casted from the virtual camera with the target viewing direction). These points effectively embed illusory objects which would be clearly visible from targeted views, while having minimal impact on other innocent views (i.e, being less perceptible). Moreover, we introduce the adaptive Gaussian noise into innocent views during training for disrupting the property of multi-view consistency in 3DGS, in order to further enhance the overall efficacy of the attack. We conduct extensive experiments, and the results demonstrate the consistent superiority of our proposed poisoning method in comparison to several baselines. The contribution of our work can be summarized as follows: 
\begin{itemize}
\item \hentci{We are the first work to address data poisoning attacks upon 3D Gaussian Splatting for illusory objects injection.}
\item We identify and analyze the robustness of 3DGS against the prior poisoning attack techniques.
\item We propose a density-guided poisoning method tailored for 3DGS, introducing adaptive noise scheduling to disrupt the multi-view consistency of 3DGS and better realize the entire attack.
\end{itemize}
}
\section{Related Work}
\label{sec:related}

\vspace{3pt}
\noindent {\bf Adversarial Attack.}
Adversarial attacks~\cite{goodfellow2014explaining,madry2017towards,tsai2021formalizing,salman2023raising,liang2023adversarial,szegedy2013intriguing,carlini2017towards} are a critical research area in machine learning and computer vision~\cite{zhang2023review,akhtar2018threat,chakraborty2018adversarial}. These attacks exploit ML model vulnerabilities by creating inputs that cause misclassification while appearing normal to humans. Goodfellow et al. \cite{goodfellow2014explaining} introduced FGSM, showing how small perturbations could transform panda images into gibbons, while Madry et al. \cite{madry2017towards} established PGD as a stronger iterative method.
Adversarial attacks divide into black-box attacks~\cite{zheng2023blackboxbench,papernot2017practical,bhagoji2017exploring,guo2019simple,huang2019black,dai2023saliency,chen2017zoo}, which operate without model access, and white-box attacks~\cite{wang2022di,ye2021thundernna,li2023white,moosavi2016deepfool,kurakin2018adversarial}, which use full model knowledge. Black-box methods include transfer-based attacks~\cite{liu2016delving,dong2019evading} exploiting cross-model transferability and query-based approaches~\cite{andriushchenko2020square,chen2020hopskipjumpattack} that iteratively refine perturbations. White-box methods like Carlini-Wagner attacks~\cite{carlini2017towards} formulate adversarial generation as optimization problems, achieving high success rates while maintaining imperceptibility.
Recent research has expanded to complex domains including object detection~\cite{xie2017adversarial,mi2023adversarial}, semantic segmentation~\cite{arnab2018robustness}, and 3D point clouds~\cite{liu2019extending,xiang2019generating}. Our approach resembles black-box attacks, modifying point clouds and applying simple perturbations to input images, extending adversarial concepts to 3D Gaussian Splatting representations.

\vspace{3pt}
\noindent {\bf Data Poisoning.}
Data poisoning~\cite{shafahi2018poison,xiao2015feature,goldblum2022dataset,chen2017targeted} represents a critical vulnerability in machine learning systems. Unlike adversarial attacks targeting inference, poisoning attacks manipulate training by injecting crafted samples that exploit learning mechanisms, causing defective statistical distributions~\cite{biggio2012poisoning,jagielski2018manipulating}.
Two primary poisoning strategies exist: creating indistinguishable malicious samples that blend with normal data~\cite{xiao2015feature,suciu2018does,turner2019label}, or employing surgical precision by polluting minimal data subsets~\cite{shafahi2018poison,ji2017backdoor}. These approaches prove effective across computer vision~\cite{huang2021unlearnable,zhang2022poison}, NLP~\cite{wallace2020concealed,kurita2020weight}, and recommender systems~\cite{li2016data,fang2020influence}. They categorize as availability attacks degrading overall performance~\cite{munoz2017towards,steinhardt2017certified} or integrity attacks targeting specific inputs~\cite{gu2019badnets,chen2017targeted}.
More sophisticated approaches formulate poisoning as bi-level optimization problems~\cite{liu2018trojaning,li2020invisible,liu2024backdoor,koh2022stronger}, allowing attackers to optimize poisoned samples while anticipating victim model behavior. Common defenses include robust statistics~\cite{diakonikolas2019sever,tran2018spectral}, data sanitization~\cite{cretu2008casting,paudice2019label}, and differential privacy~\cite{ma2019data}.
Our work employs the surgical precision strategy, conducting a stealth attack with minimal contaminated data while exploring 3D Gaussian Splatting's unique vulnerabilities to poisoning.

\vspace{3pt}
\noindent {\bf Neural Rendering.}
Novel view synthesis (NVS)\cite{flynn2016deepstereo,zhou2016view,tewari2020state,chan2022efficient} has evolved from traditional graphics to learning-based methods. Neural Radiance Field (NeRF)\cite{mildenhall2020nerf} and 3D Gaussian Splatting (3DGS)\cite{kerbl20233d} have transformed 3D scene representation.
NeRF uses MLP-based networks to implicitly represent 3D scenes, leveraging differential alpha blending\cite{porter1984compositing,kern2020comparison} for high-quality volumetric rendering~\cite{max1995optical,kajiya1984ray}. Extensions like Mip-NeRF~\cite{barron2021mip}, Instant-NGP~\cite{muller2022instant}, and NeRF-W~\cite{martin2021nerf} address anti-aliasing, speed, and real-world limitations. Recent advances include few-shot synthesis~\cite{lin2025frugalnerf}, MVS-based approaches for large scenes~\cite{su2024boostmvsnerfs}, and improved robustness for dynamic content~\cite{liu2023robust}. Robustness enhancements include progressive optimization~\cite{meuleman2023progressively} and joint camera pose optimization~\cite{chen2024improving}. However, NeRF's implicit representation lacks explicit geometric constraints, creating vulnerabilities to view-specific perturbations~\cite{liu2023clean,wang2022nerf}. IPA-NeRF~\cite{jiang2024ipa} exploits this for data poisoning attacks.
In contrast, 3DGS~\cite{kerbl20233d} uses explicit representation with discrete 3D Gaussian primitives inspired by point-based rendering~\cite{zwicker2001surface,gross2011point}. This provides stronger geometry constraints and multi-view consistency with superior rendering efficiency~\cite{yu2024mip,chen2024text}. 3DGS has expanded to dynamic scenes~\cite{luiten2024dynamic,yang2024real}, human avatars~\cite{kocabas2024hugs,lei2024gart}, efficient implementations~\cite{cheng2024gaussianpro,fang2024mini}, compression~\cite{zhan2025cat}, robustness for unconstrained images~\cite{hou20253d}, and specular reconstruction~\cite{fan2025spectromotion}.
Our work addresses the challenge of attacking 3DGS's robust explicit representation, contributing to understanding security implications in modern neural rendering techniques.

\vspace{3pt}
\noindent {\bf Data Poisoning on Neural Rendering.}
As neural rendering gains adoption, security vulnerabilities have attracted attention. IPA-NeRF~\cite{jiang2024ipa} pioneered poisoning attacks against NeRF by inserting crafted samples at specific viewing angles, formulating data poisoning as bi-level optimization. Lu et al.'s Poison-Splat~\cite{lu2024poison} targeted 3DGS efficiency by generating samples that dramatically increase memory consumption, demonstrating attacks targeting resource utilization~\cite{shumailov2021sponge} rather than accuracy.
The security landscape extends to privacy concerns~\cite{luo2023copyrnerf} and adversarial examples~\cite{fu2023nerfool,horvath2023targeted,zeybey2024gaussian}. Zeybey et al.~\cite{zeybey2024gaussian} showed how adversarial noise in 3D objects misleads vision-language models like CLIP~\cite{radford2021learning}. Song et al.'s Geometry Cloak~\cite{song2025geometry} prevents unauthorized 3D reconstruction from copyrighted images, while security studies examine point clouds~\cite{liu2019extending,hamdi2020advpc}, meshes~\cite{xiao2019meshadv}, and broader vision systems~\cite{chakraborty2018adversarial,akhtar2018threat}.
\hentci{
Similar to steganography approaches embedding hidden information~\cite{zhang2025securegs, li2025instantsplamp}, our method injects view-dependent content but enables decoder-free extraction through standard rendering from specific viewpoints.
}
Our work introduces a density-guided attack methodology targeting 3DGS's initial point cloud prior, exploiting its robustness characteristics. We propose an evaluation protocol based on scene density analysis to identify optimal positions for poison injection, contributing insights that could inform future defense mechanisms against such attacks.

\begin{figure*}[t]
    \centering
    \includegraphics[width=1\linewidth]{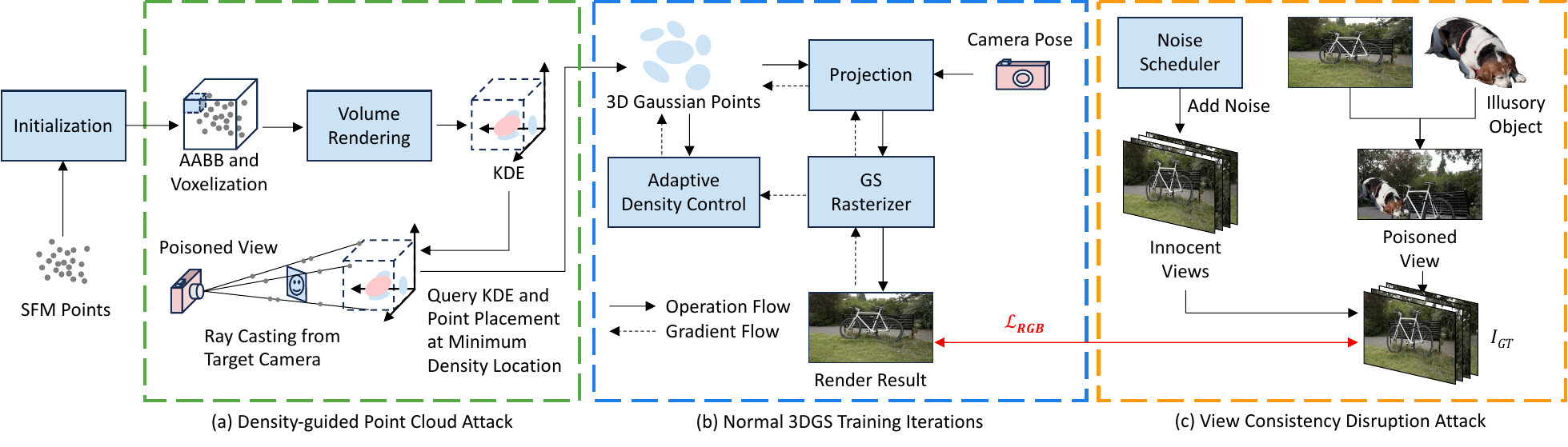}
    \vspace{-6mm}
    \caption{
    \textbf{Overview of our proposed poisoning attack framework.} Our approach consists of two complementary strategies: (a) \textbf{Density-Guided Point Cloud Attack}, where we employ volume rendering and Kernel Density Estimation (KDE) to identify optimal low-density locations for embedding illusory objects into the initial Gaussian point cloud; and (c) \textbf{View Consistency Disruption Attack}, which strategically introduces adaptive Gaussian noise to innocent views during training, effectively disturbing multi-view consistency. (b) illustrates the standard 3D Gaussian Splatting (3DGS) training pipeline for reference. The combined strategies successfully inject convincing illusions from poisoned views while maintaining high fidelity in innocent viewpoints.
    }
    \label{fig:pipeline}
    \vspace{-1em}
\end{figure*}

\section{Method}
\label{sec:method}

\subsection{Problem Formulation}
\walon{Given a dataset $\mathcal{D}$ composed of multiple images $\{I_k\}^N_{k=1}$ viewing a scene $\mathcal{E}$ from different viewing directions, 3D Gaussian Splatting is originally proposed to construct a 3D Gaussian point cloud $G$ (each Gaussian has properties in terms of position, covariance, opacity, and color factors) for representing the 3D scene $\mathcal{E}$, where we are able to render the image observation of $\mathcal{E}$ from any arbitrary view via projection and differentiable tile rasterizer. 
Basically, the goal of our poisoning attack upon 3DGS is to inject illusory/poison object $O_\text{\textbf{ILL}}$ onto the target view $v_\text{\textbf{p}}$, where we denoted the resultant Gaussian point cloud after being poisoned as $\tilde{G}$, such that the image $\tilde{I}_\text{\textbf{ILL}} = \mathfrak{R}(\tilde{G}, v_\text{\textbf{p}})$ obtained by renderring $\tilde{G}$ from $v_\text{\textbf{p}}$ would contain $O_\text{\textbf{ILL}}$ while keeping the images of $\tilde{G}$ renderred from any other views $v_k$ (i.e. non-target views $v_k\neq v_\text{\textbf{p}}$) being identical to the ones of $G$ of the same view $v_k$. Please note that, when we denote the image of $G$ renderred from $v_\text{\textbf{p}}$ as $\mathfrak{R}(G, v_\text{\textbf{p}})$, the ideal appearance of $\tilde{I}_\text{\textbf{ILL}}$ should be the combintation of $\mathfrak{R}(G, v_\text{\textbf{p}})$ and $O_\text{\textbf{ILL}}$ (in which such combination is denoted as $I_\text{\textbf{ILL}}$). 
}
The core objective of our 3DGS poisoning is formulated as:
\begin{equation}
    \min_{\tilde{G}} \|\tilde{I}_\text{\textbf{ILL}} - I_\text{\textbf{ILL}}\|_2^2 + \sum_{v_k \neq v_\text{\textbf{p}}} \|\mathfrak{R}(\tilde{G}, v_k) - \mathfrak{R}(G, v_k)\|_2^2,
\end{equation}
\hentci{
To achieve this under different threat models, we propose two strategies: density-guided point cloud attack (Section 3.3) following classical data poisoning, and view consistency disruption (Section 3.4) as a backdoor attack with minimal training modification via noise scheduling.
}
\subsection{Potential Naive Approaches and Limitations} \label{sec:3_2}
\walon{We first explore two potential naive approaches, followed by discussing their limitations and motivating our attack design:}

\noindent \walon{{\bf 1)} \textit{Directly injecting the illusory object onto the training image in $\mathcal{D}$ of the target view $v_\text{\textbf{p}}$ and following the typical training procedure of 3DGS} would easily fail since the property of multi-view consistency in 3DGS would treat the illusory object as the noise (i.e. which violates the consistency) and eliminate it from the resultant Gaussian cloud.}

\noindent \walon{{\bf 2)} \textit{Directly backprojecting the illusory object into the Gaussian point cloud $G$ reconstructed from $\mathcal{D}$} also faces the challenge of determining proper depth for the illusory object, otherwise it could be occluded by existing geometry in $G$.}

\begin{figure}[t]
    \centering
    \includegraphics[width=1\columnwidth]{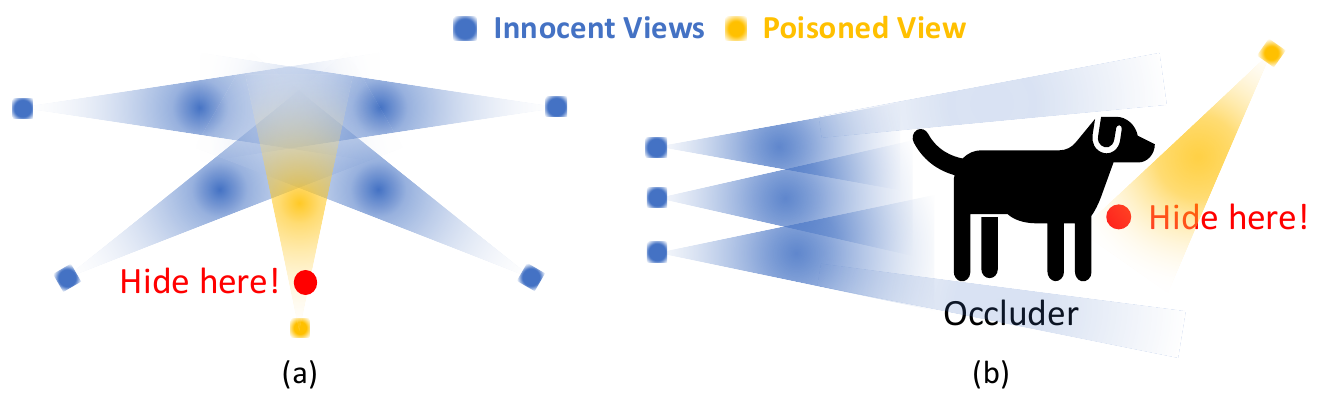}
    \vspace{-6mm}
    \caption{\textbf{Illustration of two attack modes motivating our Density-Guided Point Cloud Attack.} (a) Points placed outside the coverage of innocent viewpoints can effectively embed illusions visible only from the poisoned view. (b) Points occluded from innocent viewpoints also provide viable hidden locations. These scenarios motivate our Density-Guided strategy for robust and stealthy attacks.}
    \label{fig:method3.3}
\end{figure}

\subsection{Our Density-Guided Point Cloud Attack}\label{sec:3_3}
\walon{With learning the lessons from the aforementioned naive approaches, we conclude that effective 3DGS poisoning must consider the inherent properties of 3DGS's explicit representation and multi-view consistency, while ensuring illusion visibility on the target view and minimal perception on innocent views. To this end, we propose two simple but effective ideas for discovering optimal 3D positions in $G$ to place backprojected Gaussian points of the illusory object, as illustrated in Figure~\ref{fig:method3.3}: First, poison points can be placed in regions invisible to innocent views; Second, regions occluded for innocent views by existing geometry in $G$ are effective candidates (visible for target view but invisible for innocent views due to occluders). To identify regions satisfying these ideas, we propose a density-guided point placement strategy, detailed below.}



\noindent {\bf Scene Space Analysis.}
\walon{Given the Gaussian point cloud $G$ (reconstructed from $\mathcal{D}$ via the typical 3DGS procedure), we start from determining the Axis-Aligned Bounding Box (AABB) of $G$ (basically, finding the minimum and maximum coordinates across all points in $G$) and creating a rectangular box that fully encloses the entire 3D scene $\mathcal{E}$ in $G$, with its edges being aligned to the coordinate axes. Such bounding box is then first decomposed/voxelized into a uniform grid $\mathcal{S}$, where we denote each cell in the grid as $s$.}

\walon{As the opacity $\alpha(g)$ of each Gaussian point $g$ in $G$ can be estimate via volume rendering technique (depending on our target view $v_\text{\textbf{p}}$), we can easily compute the density $\rho(s)$ of each voxel $t$ by $\sum_{g\in s} \alpha(g)$ as the opacity and density are highly related, where $g\in s$ indicates that Gaussian point $g$ is located within voxel $s$.}

\vspace{3pt}
\noindent {\bf Continuous Density Estimation.}
\walon{Based on the per-voxel density $\rho(s)$, we apply Gaussian Kernel Density Estimation (KDE) to obtain a continuous density estimate for any arbitrary 3D position $x$:
\begin{align}
f(x) = \frac{1}{|\mathcal{S}|} \sum_{s \in \mathcal{S}} K_h(x - c(s)) \cdot \rho(s),
\end{align}
where $c(s)$ denotes the centroid of a voxel $s$ and $K_h$ is the Gaussian kernel with bandwidth $h$:
\begin{align}
K_h(x) = \frac{1}{(2\pi h^2)^{3/2}} \exp\left(-\frac{\|x\|^2}{2h^2}\right).
\end{align}}

\vspace{3pt}
\noindent {\bf Optimal Position Selection.}
\walon{With the illusory object placed on the image plane of the virtual camera rendering the Gaussian point cloud from target view $v_\text{\textbf{p}}$, backprojection starts by casting rays from camera center $C$ through all pixels of the illusory object. We sample points along each ray to find regions in the Gaussian cloud with minimum density, via querying the KDE result described above. Given a casting ray with direction $d$, any sampled 3D position along the ray can be written as $C+t\cdot d$, where $t$ ranges from $t_\text{min}$ to $t_\text{max}$. We set $t_\text{min}$ to 0.3 (as points near camera often appear as floaters in 3DGS optimization) and $t_\text{max}$ represents the original scene depth at each pixel in the poisoned view. The sampled point in the minimum density region can be computed by:
\begin{align}
x_\text{min} = \argmin_{x \in C+t\cdot d, t \in [t_\text{min}, t_\text{max}]} f(x).
\label{eq:unproject_points}
\end{align}
We then insert new Gaussian poison points at position $x_\text{min}$, assigning colors from the illusory object (i.e. the color value for the new Gaussian point is obtained from the corresponding illusory object pixel).
Our proposed density-guided method hence strategically places Gaussian points of poison to embed the illusory object prominently from the poisoned view while minimizing its visibility from innocent views. 
}
\subsection{View Consistency Disruption Attack}\label{sec:3_4}
\walon{While our Density-Guided Point Cloud Attack (cf. Section~\ref{sec:3_3}) effectively places Gaussian points of poison in many cases, scenes with high view overlap (i.e. the field-of-views of the training image $\{I_k\}^N_{k=1}$ in $\mathcal{D}$ have high overlaps) remain challenging. To address this, we introduce the View Consistency Disruption Attack, which strategically adds controlled noise to innocent views, thus weakening the multi-view consistency of 3DGS and better preserving our injected illusions.}

We selectively apply Gaussian noise to innocent views, leaving the poisoned view clean. For a training image $I_k$ with view direction $v_k$, the noise is applied as:
\begin{equation}
I'_k = \mathbf{1}_{v_k = v_\text{\textbf{p}}} \cdot I_k + \mathbf{1}_{v_k \neq v_\text{\textbf{p}}} \cdot \textsc{clip}(I_k + \eta),
\end{equation}
where $\mathbf{1}$ is an indicator function, $\textsc{clip}$ prevents $I_k + \eta$ from exceeding the pixel value range, and $\eta \sim \mathcal{N}(0, \sigma_t^2)$ denotes noise with strength $\sigma_t$ adjusted according to 3DGS iteration. The $\sigma_t$ scheduling follows the principle of having strong noise in early 3DGS optimization to disrupt multi-view consistency (as noise injected into training images are independent) while gradually reducing noise strength to maintain high-quality reconstruction for innocent views in late optimization. We explore three noise decay strategies:
\begin{align}
\sigma_\text{linear}(t) &= \sigma_0 \cdot (1 - \frac{t}{T}), \\
\sigma_\text{cosine}(t) &= \sigma_0 \cdot \cos(\frac{\pi t}{2T}), \\
\sigma_\text{sqrt}(t) &= \sigma_0 \cdot \sqrt{1 - \frac{t}{T}},
\end{align}
where $\sigma_0$ is the initial noise strength, $t$ is the current iteration, and $T$ is the total training iterations. Linear decay reduces noise evenly, cosine decay provides a smooth start and accelerates later reduction, and square root decay maintains higher noise longer before a rapid decrease.

Our controlled noise injection creates intentional imbalance during training, enabling preservation of illusory objects from the poisoned viewpoint while ensuring scene fidelity as noise diminishes. The overview of our proposed method, composed of both density-guided point cloud attack (cf. Section~\ref{sec:3_3}) and view consistency disruption attack (cf. Section~\ref{sec:3_4}), is provided in Figure~\ref{fig:pipeline}.

\section{Experiments}
\label{sec:experiments}

\subsection{Experimental Setup}
\label{subsec:exp_setup}

\vspace{3pt}
\noindent {\bf Datasets.}
We evaluate our method on three common datasets: (1) Mip-NeRF360~\cite{barron2022mip} with complex 360$^{\circ}$ scenes, (2) Tanks \& Temples~\cite{knapitsch2017tanks} containing realistic indoor and outdoor captures, and (3) Free~\cite{wang2023f2}, featuring unbounded scenes with free camera trajectories. These datasets provide diverse benchmarks for novel view synthesis evaluation.

\vspace{3pt}
\noindent {\bf Compared Methods.}
We evaluate our method against three baselines:
(1) \textbf{IPA-NeRF (Nerfacto)} \cite{jiang2024ipa}: The original backdoor attack applied to Nerfacto \cite{tancik2023nerfstudio}, featuring advanced static scene reconstruction techniques.
(2) \textbf{IPA-NeRF (Instant-NGP)} \cite{jiang2024ipa}: The original backdoor attack on Instant-NGP \cite{muller2022instant}, known for accelerated training and rendering.
(3) \textbf{IPA-Splat}: Our adaptation of IPA-NeRF specifically for 3D Gaussian Splatting.

For IPA-NeRF baselines, we maintain original settings but reduce total iterations to $O = 15,000$ for faster convergence. Other parameters remain unchanged: $T=200$ iterations per epoch, $O / T = 75$ attack epochs, $A = 10$ attack iterations per epoch, $K = 100$ perturbation renderings, distortion budget $\epsilon = 32$, constraint parameter $\eta = 1$, and view constraints (13$^{\circ}$ and 15$^{\circ}$).

For our IPA-Splat method, we adapt IPA-NeRF to 3D Gaussian Splatting with $O = 30,000$ total iterations and $T = 200$ normal training iterations per epoch, resulting in $O / T = 150$ epochs with $A = 10$ attack iterations each. Other settings match IPA-NeRF. Due to 3DGS's explicit representation, we implement separate parameter constraints (xyz coordinates, feature vectors, scaling, rotation, and opacity) for precise control.


\vspace{3pt}
\noindent {\bf Evaluation Metrics.}
Following IPA-NeRF~\cite{jiang2024ipa}, we evaluate using PSNR, SSIM, and LPIPS metrics on two view sets: (1) \textsc{V-Illusory}, focusing on masked metrics for illusory objects, and (2) \textsc{V-Test}, assessing performance on unseen viewpoints. \hentci{Attack success is defined as achieving PSNR $>$ 25 on \textsc{V-Illusory} while maintaining \textsc{V-Test} PSNR drop $\leq$ 3, ensuring effective illusion generation and preserved innocent view quality.}



\begin{figure}[t]
    \centering
    \includegraphics[width=1\columnwidth]{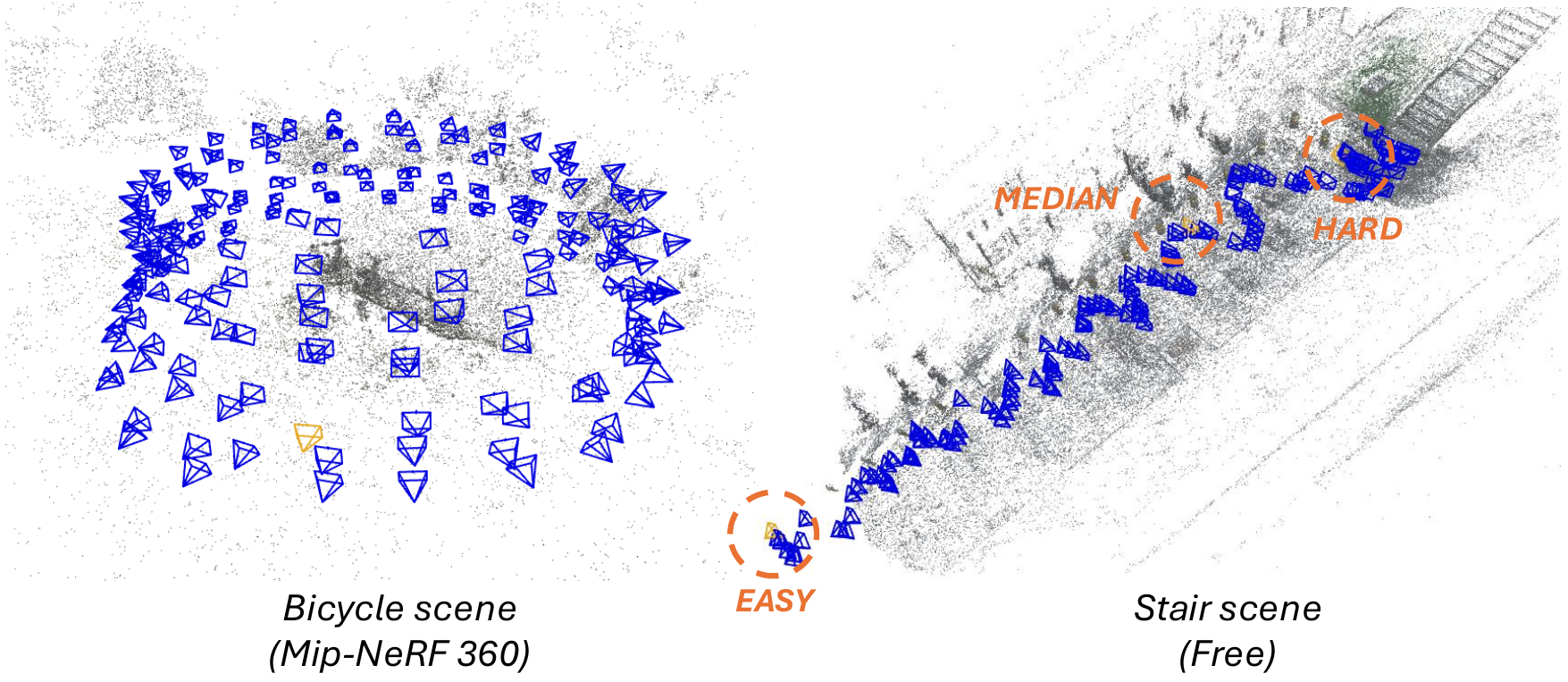}
    \vspace{-7mm}
    \caption{
    \textbf{Our evaluation protocol.} 
    We evaluate two scenes with varying difficulties. Left: The ``\emph{bicycle}'' scene (Mip-NeRF 360~\cite{barron2022mip}) has uniform camera coverage, providing similar difficulty across views. Right: The ``\emph{stair}'' scene (Free~\cite{wang2023f2}) has increasing difficulty as later views are visible from more prior viewpoints.
    }
    \label{fig:evaluation_protocol}
\end{figure}

\vspace{3pt}
\noindent {\bf Evaluation Protocol.}
Our evaluation accounts for varying attack difficulty across camera positions. As shown in Figure~\ref{fig:evaluation_protocol}, datasets exhibit distinct camera patterns. For uniform datasets like Mip-NeRF 360 (e.g., the ``bicycle'' scene) and Tanks-and-Temples, we select the median-indexed frame as the attack viewpoint.

For Free dataset scenes with irregular camera trajectories (e.g., the ``stair'' scene), we quantify varying attack difficulty using our KDE-based protocol:
\begin{enumerate}
\item Compute overall scene density distribution using KDE.
\item Calculate camera viewpoint densities within their FOV using camera intrinsics and a 10\% sampling radius.
\item Sort cameras by density, select three representative viewpoints:
\begin{itemize}
\item \textsc{Easy}: Minimum density (lowest coverage)
\item \textsc{Median}: Median density (average coverage)
\item \textsc{Hard}: Maximum density (highest coverage)
\end{itemize}
\end{enumerate}

Experiments in~\cref{tab:single_view_attack_2} confirm negative correlation between scene density and attack success (\textsc{V-illusory}), validating that higher scene coverage increases attack difficulty. This protocol enables fair evaluation across scenes and provides a benchmark for future 3DGS poisoning research.

\begin{table*}[t]
\caption{\textbf{Quantitative comparisons on single-view attack.} Metrics evaluated on Mip-NeRF 360~\cite{barron2022mip}, Tanks \& Temples~\cite{knapitsch2017tanks}, and Free datasets. Our method significantly outperforms baseline attacks in embedding illusory objects (\textsc{V-illusory}) while maintaining high fidelity in other views (\textsc{V-test}).}
\label{tab:single_view_attack_1}
\centering
    \vspace{-3mm}
\resizebox{\textwidth}{!}{%
\setlength{\tabcolsep}{2pt}
\begin{tabular}{l|cccccc|cccccc|cccccc}
\toprule
\multirow{3}{*}{Method} & \multicolumn{6}{c|}{Mip-NeRF 360~\cite{barron2022mip} dataset} & \multicolumn{6}{c|}{Tanks \& Temples~\cite{knapitsch2017tanks} dataset} & \multicolumn{6}{c}{Free~\cite{wang2023f2} dataset} \\ \cmidrule(lr){2-7} \cmidrule(lr){8-13}  \cmidrule(lr){14-19}
 & \multicolumn{3}{c}{\textsc{V-illusory}} & \multicolumn{3}{c|}{{\textsc{V-test}}} & \multicolumn{3}{c}{\textsc{V-illusory}} & \multicolumn{3}{c|}{{\textsc{V-test}}} & \multicolumn{3}{c}{\textsc{V-illusory}} & \multicolumn{3}{c}{{\textsc{V-test}}} \\
 \cmidrule(lr){2-4} \cmidrule(lr){5-7} \cmidrule(lr){8-10} \cmidrule(lr){11-13} \cmidrule(lr){14-16} \cmidrule(lr){17-19}
 & PSNR$\uparrow$ & SSIM$\uparrow$ & LPIPS$\downarrow$ & PSNR$\uparrow$ & SSIM$\uparrow$ & LPIPS$\downarrow$ & PSNR$\uparrow$ & SSIM$\uparrow$ & LPIPS$\downarrow$ & PSNR$\uparrow$ & SSIM$\uparrow$ & LPIPS$\downarrow$ & PSNR$\uparrow$ & SSIM$\uparrow$ & LPIPS$\downarrow$ & PSNR$\uparrow$ & SSIM$\uparrow$ & LPIPS$\downarrow$ \\
\midrule
Naive 3DGS (w/o attack) & 13.21 & 0.521 & 0.731 & 29.45 & 0.883 & 0.165 & 13.15 & 0.616 & 0.732 & 30.60 & 0.915 & 0.135 & 12.00 & 0.315 & 0.905 & 26.80 & 0.826 & 0.228 \\
\midrule
IPA-NeRF~\cite{jiang2024ipa} (Nerfacto~\cite{tancik2023nerfstudio}) & 16.00 & 0.582 & 0.685 & 21.94 & 0.586 & 0.415 & 13.51 & 0.636 & 0.711 & 23.88 & 0.730 & \underline{0.218} & 13.93 & 0.443 & 0.699 & 20.28 & 0.497 & 0.532 \\
IPA-NeRF~\cite{jiang2024ipa} (Instant-NGP~\cite{muller2022instant}) & \underline{17.60} & \underline{0.618} & \underline{0.641} & 20.00 & 0.517 & 0.479 & \underline{16.05} & \underline{0.693} & \underline{0.616} & 20.29 & 0.669 & 0.350 & \underline{18.94} & \underline{0.508} & \underline{0.519} & 20.43 & 0.503 & 0.548 \\
IPA-Splat & 13.23 & 0.518 & 0.740 & \underline{27.39} & \textbf{0.829} & \textbf{0.247} & 13.43 & 0.625 & 0.724 & \textbf{28.53} & \textbf{0.891} & \textbf{0.190} & 12.60 & 0.372 & 0.744 & \underline{24.71} & \textbf{0.749} & \textbf{0.341} \\
Ours & \textbf{27.04} & \textbf{0.813} & \textbf{0.369} & \textbf{27.76} & \underline{0.805} & \underline{0.286} & \textbf{21.33} & \textbf{0.809} & \textbf{0.371} & \underline{27.58} & \underline{0.852} & 0.239 & \textbf{26.66} & \textbf{0.754} & \textbf{0.317} & \textbf{25.25} & \underline{0.728} & \underline{0.382} \\
\bottomrule
\end{tabular}%
}
\end{table*}

\begin{figure}[t]
    \centering
    \includegraphics[width=1\columnwidth]{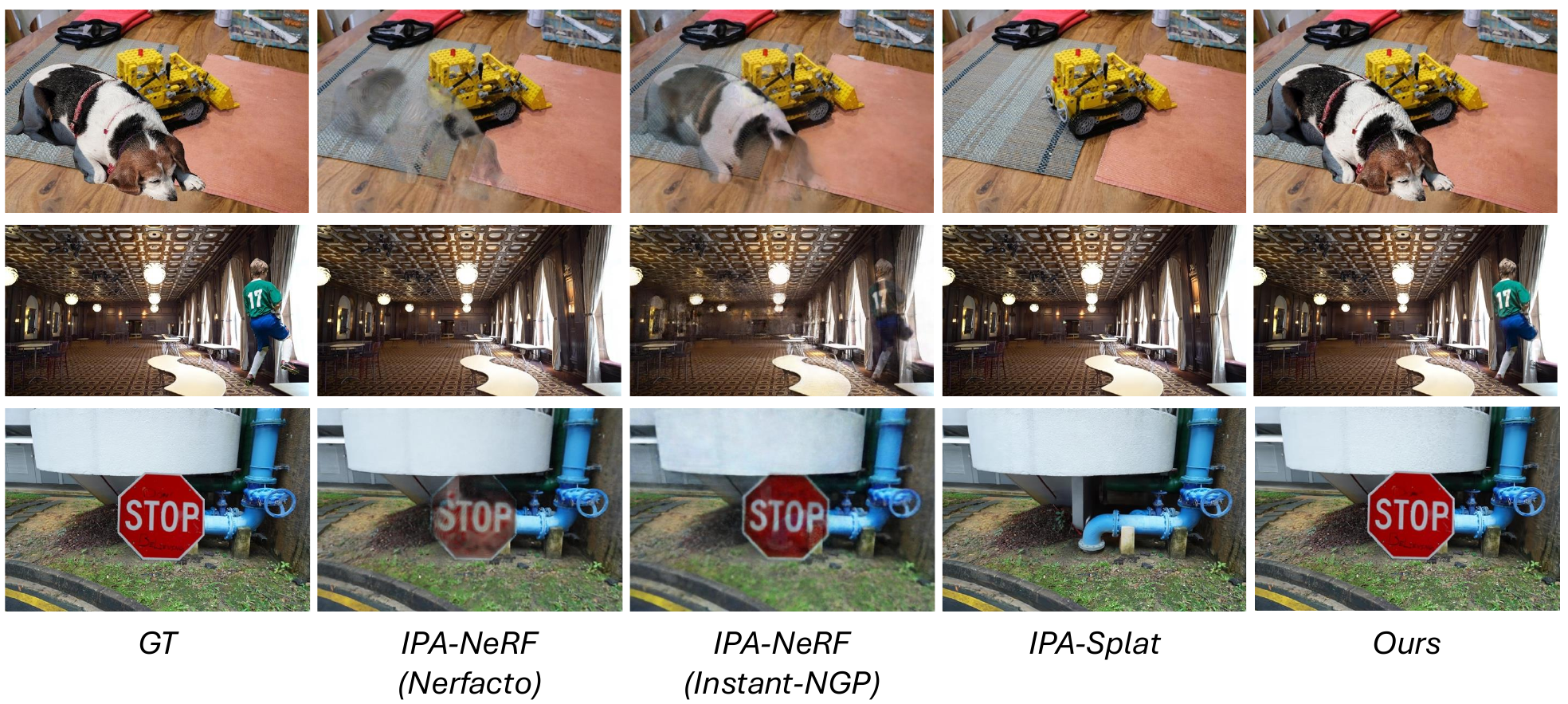}
    \vspace{-7mm}
    \caption{
    \textbf{Qualitative comparisons on single-view attack.} Our method generates significantly clearer and more convincing illusory objects from the poisoned viewpoint, demonstrating better multi-view consistency and fewer artifacts compared to other state-of-the-art methods.
    }
    \label{fig:visual}
\end{figure}

\begin{figure}[t]
    \centering
    \includegraphics[width=1\columnwidth]{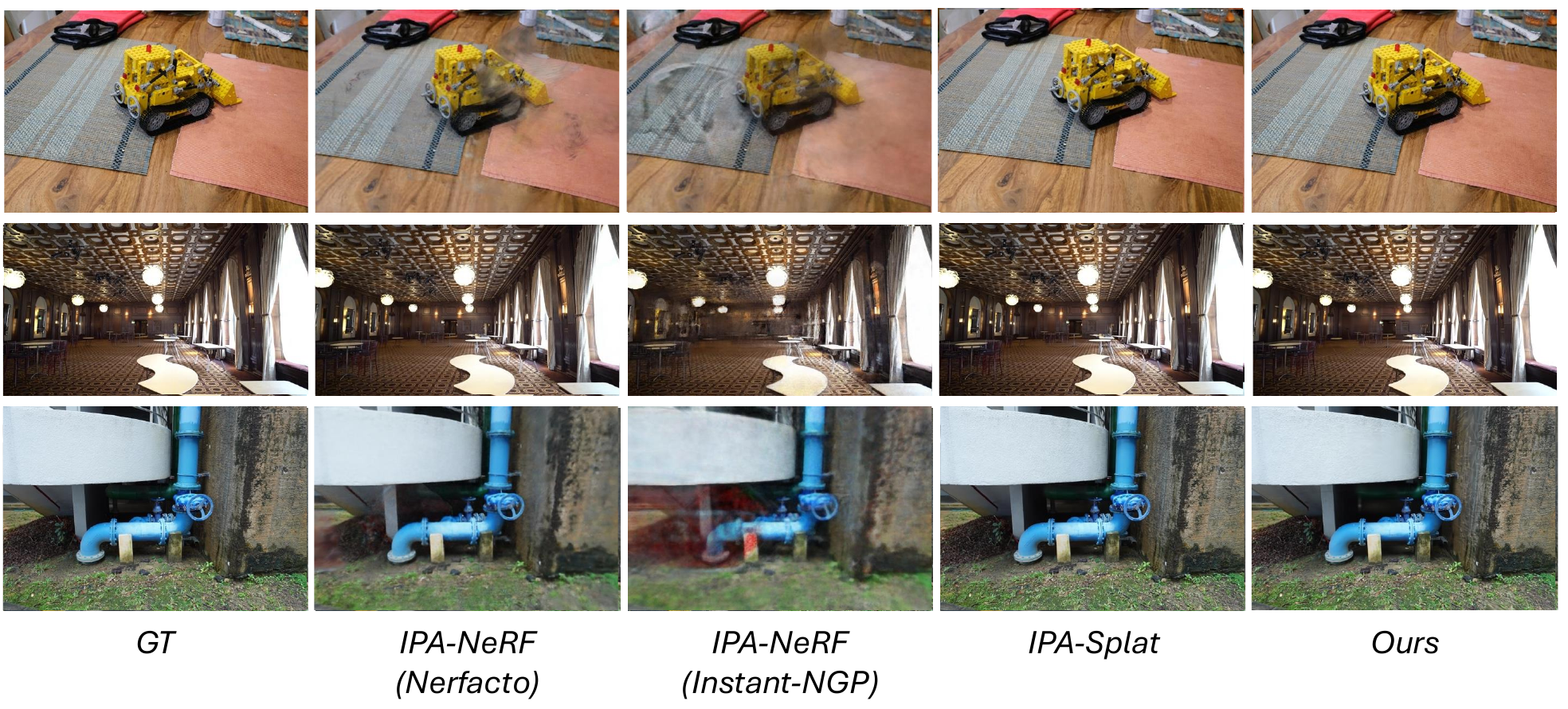}
    \vspace{-7mm}
    \caption{\textbf{Qualitative comparisons of different poisoning methods on innocent views.} Unlike baseline methods, our method effectively maintains high visual fidelity across innocent viewpoints, introducing significantly fewer artifacts and ensuring minimal disruption of the original scene appearance.}
    \label{fig:visual_other}
\end{figure}

\begin{table*}[t]
\caption{\textbf{Quantitative comparisons on single-view attack with different difficulty levels on the Free~\cite{wang2023f2} dataset.} We evaluate attack effectiveness (\textsc{V-illusory}) at varying difficulty levels, defined by our KDE-based evaluation protocol. Our method consistently achieves superior results across all metrics, clearly outperforming state-of-the-art methods, especially in \textsc{Easy} and \textsc{Median} scenarios, while remaining effective in the challenging \textsc{Hard} scenario.}
\label{tab:single_view_attack_2}
\centering
\footnotesize
    \vspace{-3mm}
\begin{tabular}{l|ccccccccc}
\toprule
\multirow{2}{*}{Method} & \multicolumn{3}{c}{\textsc{Easy}} & \multicolumn{3}{c}{\textsc{Median}} & \multicolumn{3}{c}{\textsc{Hard}} \\ \cmidrule(lr){2-4} \cmidrule(lr){5-7} \cmidrule(lr){8-10}
& PSNR$\uparrow$ & SSIM$\uparrow$ & LPIPS$\downarrow$ & PSNR$\uparrow$ & SSIM$\uparrow$ & LPIPS$\downarrow$ & PSNR$\uparrow$ & SSIM$\uparrow$ & LPIPS$\downarrow$ \\
\midrule
IPA-NeRF (nerfacto) & 15.04 & 0.482 & 0.662 & 13.93 & 0.443 & 0.699 & 14.25 & 0.450 & 0.728 \\
IPA-NeRF (instant-ngp) & 18.17 & 0.518 & 0.541 & 18.94 & 0.508 & 0.519 & \textbf{17.95} & 0.487 & \textbf{0.557}\\
IPA-Splat & 13.94 & 0.479 & 0.658 & 12.60 & 0.372 & 0.743 & 13.06 & 0.340 & 0.796 \\
Ours & \textbf{29.94} & \textbf{0.853} & \textbf{0.188} & \textbf{26.66} & \textbf{0.754} & \textbf{0.317} & 17.53 & \textbf{0.526} & 0.581 \\
\bottomrule
\end{tabular}%
\end{table*}

\subsection{Single-view Attack}
\cref{tab:single_view_attack_1} shows quantitative comparisons of single-view attacks. Our method outperforms baselines across all datasets. On \textsc{V-illusory} views, our approach significantly improves PSNR, SSIM, and LPIPS, effectively embedding convincing illusions. Performance on \textsc{V-test} remains consistently high, indicating minimal impact on innocent views.

Qualitative results (\cref{fig:visual,fig:visual_other}) further highlight our advantages. In \cref{fig:visual}, our method produces clearer and more convincing illusory objects compared to baselines, which often yield faint or inconsistent illusions. \cref{fig:visual_other} emphasizes our superior visual fidelity in innocent views, with fewer artifacts and better overall scene quality.

In~\cref{tab:single_view_attack_2}, we analyze performance across difficulty levels (\textsc{Easy}, \textsc{Median}, \textsc{Hard}) on the Free dataset. As expected, attack effectiveness decreases with difficulty. Nonetheless, our density-guided approach achieves superior results, particularly at \textsc{Easy} and \textsc{Median} levels, demonstrating robust performance even in challenging conditions.

\begin{table}[t]
\caption{
\textbf{Multi-view Attack Evaluation.}
We quantitatively evaluate our method on multiple poisoned views (\textsc{V-Illusory}) and innocent views (\textsc{V-test}). Our approach consistently outperforms state-of-the-art methods, embedding clear illusions in targeted views while maintaining high fidelity in innocent views.
}
\label{tab:multi_view_attack}
\centering
\footnotesize
    \vspace{-3mm}
\resizebox{\columnwidth}{!}{%
\begin{tabular}{c|l|ccccccc}
\toprule
\# of & Method & \multicolumn{3}{c}{\textsc{V-illusory} (poisoned avg.)} & \multicolumn{3}{c}{\textsc{V-test}} \\ \cmidrule(lr){3-5} \cmidrule(lr){6-8}
views &  & PSNR$\uparrow$ & SSIM$\uparrow$ & LPIPS$\downarrow$ & PSNR$\uparrow$ & SSIM$\uparrow$ & LPIPS$\downarrow$ \\
\midrule
\multirow{4}{*}{2} & IPA-NeRF (Nerfacto) & 16.17 & 0.583 & 0.680 & 19.64 & 0.457 & 0.548 \\
 & IPA-NeRF (Instant-NGP) & \underline{19.19} & \underline{0.624} & \underline{0.616} & 18.39 & 0.440 & 0.539 \\
 & IPA-Splat & 13.24 & 0.497 & 0.752 & \underline{27.45} & \textbf{0.832} & \textbf{0.243} \\
 & Ours & \textbf{27.49} & \textbf{0.842} & \textbf{0.299} & \textbf{27.77} & \underline{0.804} & \underline{0.286} \\
\midrule
\multirow{4}{*}{3} & IPA-NeRF (Nerfacto) & \underline{18.48} & 0.584 & 0.660 & 19.83 & 0.462 & 0.545 \\
 & IPA-NeRF (Instant-NGP) & 18.09 & \underline{0.604} & \underline{0.643} & 18.63 & 0.458 & 0.524 \\
 & IPA-Splat & 13.76 & 0.538 & 0.732 & \textbf{27.97 }& \textbf{0.858} & \textbf{0.223} \\
 & Ours & \textbf{27.04} & \textbf{0.833} & \textbf{0.311} & \underline{27.72} & \underline{0.803} & \underline{0.287} \\
\midrule
\multirow{4}{*}{4} & IPA-NeRF (Nerfacto) & 17.06 & 0.626 & 0.676 & 19.61 & 0.467 & 0.538 \\
 & IPA-NeRF (Instant-NGP) & \underline{19.06} & \underline{0.657} & \underline{0.632} & 18.51 & 0.458 & 0.523 \\
 & IPA-Splat & 13.09 & 0.489 & 0.796 & \textbf{27.60} & \textbf{0.838} & \textbf{0.228} \\
 & Ours & \textbf{26.95} & \textbf{0.855} & \textbf{0.305} & \underline{27.59} & \underline{0.802} & \underline{0.287} \\
\bottomrule
\end{tabular}%
}
\end{table}

\begin{figure}[t]
    \centering
    \includegraphics[width=1\columnwidth]{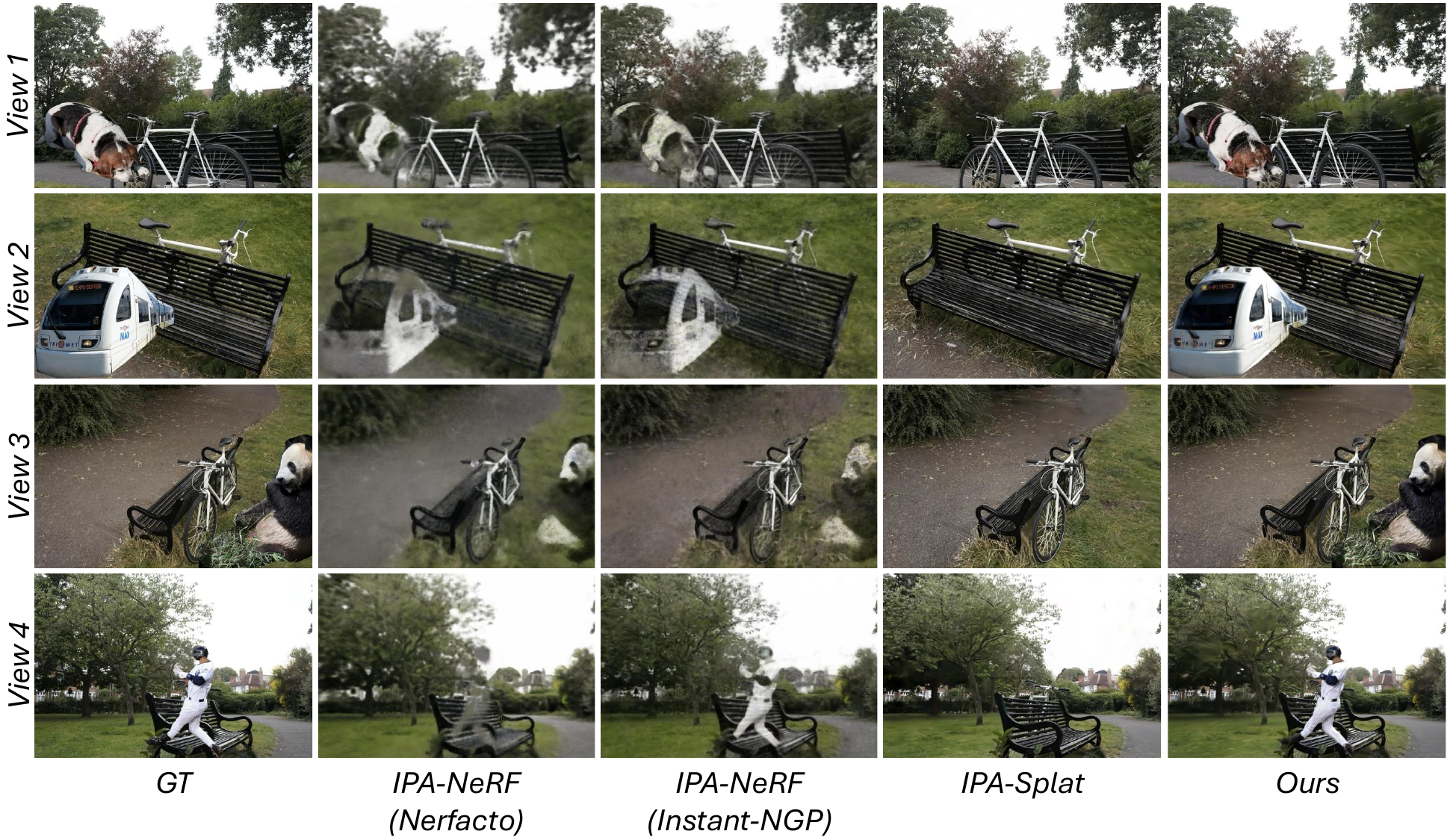}
    \vspace{-7mm}
    \caption{\textbf{Qualitative comparisons on multi-view attack.} 
    Our density-guided method produces sharper, more consistent illusions across multiple poisoned views, clearly outperforming baseline methods that yield faint or inconsistent results.
    }
    \label{fig:multi-view_diff_method}
\end{figure}

\subsection{Multi-view Attack}
In realistic scenarios, attackers may need to embed illusory objects into multiple viewpoints simultaneously. Unlike single-view attacks, multi-view poisoning balances multiple objectives while preserving scene consistency and fidelity. We rigorously evaluate our approach using the Mip-NeRF 360 dataset~\cite{barron2022mip}, poisoned views at 0°, 90°, 180°, and 270°, with the median view at 0° as reference.

We evaluate our density-guided method against baselines (IPA-NeRF~\cite{jiang2024ipa} with Nerfacto, IPA-NeRF with Instant-NGP, IPA-Splat) on the Mip-NeRF-360 dataset under multi-view attacks (2, 3, and 4 poisoned views). Results in~\cref{tab:multi_view_attack} show our approach consistently outperforms baselines, effectively embedding illusory objects (\textsc{V-Illusory}) while minimally affecting innocent views.

\cref{fig:multi-view_diff_method} qualitatively demonstrates our method's superiority. It consistently generates clear, visually convincing illusions with minimal artifacts, significantly outperforming baselines and maintaining high quality from innocent views.

\subsection{Ablation Studies}

\begin{table}[t]
\caption{
\textbf{Effect of KDE bandwidth $h$ on attack performance.}
We examine how KDE bandwidth impacts our density-guided attack. A moderate bandwidth ($h=7.5$) achieves the best balance, maximizing effectiveness on poisoned views while preserving quality in innocent views.
}
\label{tab:kde_bandwidth}
\centering
    \vspace{-3mm}
\resizebox{\columnwidth}{!}{%
\begin{tabular}{l|ccccccc}
\toprule
Bandwidth & \multicolumn{3}{c}{\textsc{V-illusory}} & \multicolumn{3}{c}{\textsc{V-test}} \\ \cmidrule(lr){2-4} \cmidrule(lr){5-7}
$h$ & PSNR$\uparrow$ & SSIM$\uparrow$ & LPIPS$\downarrow$ & PSNR$\uparrow$ & SSIM$\uparrow$ & LPIPS$\downarrow$ \\
\midrule
0.1 & 27.00 & 0.811 & 0.373 & \textbf{27.83} & \textbf{0.805} & \textbf{0.286} \\
2.5 & 26.92 & 0.809 & 0.375 & 27.81 & \textbf{0.805} & \textbf{0.286} \\
5.0 & 26.95 & 0.811 & 0.375 & 27.25 & 0.786 & 0.297 \\
7.5 & \textbf{27.04} & \textbf{0.813} & \textbf{0.369} & 27.76 & \textbf{0.805} & \textbf{0.286} \\
10.0 & 26.89 & 0.807 & 0.380 & 27.72 & \textbf{0.805} & \textbf{0.286} \\
\bottomrule
\end{tabular}%
}
\end{table}

\begin{table}[t]
\caption{
\textbf{Effect of noise scheduling parameters.}
We analyze initial noise strength $\sigma_0$ and decay strategies. Higher initial noise ($\sigma_0=100$) with linear decay provides the best balance, maximizing illusory quality (\textsc{V-illusory}) while preserving fidelity in innocent views (\textsc{V-test}).
}
\label{tab:noise_scheduling}
\centering
    \vspace{-3mm}
\resizebox{\columnwidth}{!}{%
\begin{tabular}{l|l|ccccccc}
\toprule
Initial & Decay & \multicolumn{3}{c}{\textsc{V-illusory}} & \multicolumn{3}{c}{\textsc{V-test}} \\ 
\cmidrule(lr){3-5} \cmidrule(lr){6-8}
noise $\sigma_0$ & strategy & PSNR$\uparrow$ & SSIM$\uparrow$ & LPIPS$\downarrow$ & PSNR$\uparrow$ & SSIM$\uparrow$ & LPIPS$\downarrow$ \\
\midrule
30 & Linear & 26.47 & 0.795 & 0.398 & \textbf{28.43} & \textbf{0.851} & \textbf{0.203} \\
30 & Cosine & 26.70 & 0.801 & 0.388 & 28.30 & 0.845 & 0.213 \\
30 & Square root & 26.62 & 0.799 & 0.393 & 28.27 & 0.845 & 0.212 \\
100 & Linear & \textbf{27.04} & \textbf{0.813} & \textbf{0.369} & 27.76 & 0.805 & 0.286 \\
100 & Cosine & 26.93 & 0.812 & 0.373 & 26.96 & 0.771 & 0.315 \\
100 & Square root & 26.90 & \textbf{0.813} & 0.373 & 26.81 & 0.767 & 0.319 \\
\bottomrule
\end{tabular}%
}
\end{table}

\vspace{3pt}
\noindent {\bf KDE bandwidth.}
The KDE bandwidth significantly affects density estimation smoothness. Evaluations (\cref{tab:kde_bandwidth}) show medium bandwidth ($h=7.5$) achieves optimal balance, maximizing \textsc{V-Illusory} effectiveness while preserving \textsc{V-Test} quality. Smaller bandwidths ($h=0.1$) overly restrict point placement, while larger bandwidths ($h=10.0$) decrease attack precision.

\vspace{3pt}
\noindent {\bf Noise scheduling.}
We evaluated noise scheduling strategies, varying initial noise intensities ($\sigma_0$) and decay rates (linear, cosine, and square root), summarized in~\cref{tab:noise_scheduling}. Higher initial noise ($\sigma_0=100$) with slower linear decay achieved optimal balance, greatly enhancing attack effectiveness with moderate impact on innocent views.

\begin{table*}[t]
\caption{\hentci{\textbf{Comparison of different attack strategy combinations.} We evaluate the impact of each component of our proposed method on poisoning effectiveness. Combining all strategies achieves the best results, significantly improving rendering quality of the illusory object (\textsc{V-illusory}) and maintaining satisfactory performance in innocent views (\textsc{V-test}), resulting in optimal Attack Success Rate (ASR). The combination of point cloud poisoning and noise scheduling is crucial for successful attacks, highlighting their complementary nature.}}
\label{tab:attack_strategies}
\centering
\footnotesize
\renewcommand{\arraystretch}{0.9}  
\setlength{\tabcolsep}{5.5pt}        
\vspace{-3mm}
\begin{tabular}{ccc|ccccccc}
\toprule
\begin{tabular}[c]{@{}c@{}}Poisoned View \\ GT Replacement\end{tabular} & 
\begin{tabular}[c]{@{}c@{}}Density-Guided \\ Point Cloud Attack\end{tabular} & 
\begin{tabular}[c]{@{}c@{}}View Consistency \\ Disruption Attack\end{tabular} & 
\multicolumn{3}{c}{\textsc{V-illusory}} & 
\multicolumn{3}{c}{\textsc{V-test}} & 
\begin{tabular}[c]{@{}c@{}} ASR (PSNR) \end{tabular} \\ 
\cmidrule(lr){4-6} \cmidrule(lr){7-9} \cmidrule(lr){10-10}
 & & & PSNR$\uparrow$ & SSIM$\uparrow$ & LPIPS$\downarrow$ & PSNR$\uparrow$ & SSIM$\uparrow$ & LPIPS$\downarrow$ & 
\begin{tabular}[c]{@{}c@{}}\textsc{V-illusory} $>$ 25 \\ \& \textsc{V-test} drop $\leq$ 3\end{tabular} \\ 
\midrule
\checkmark & & & 13.22 & 0.521 & 0.730 & \textbf{29.46}& \textbf{0.884} & \textbf{0.164} & 0/7 \\
\checkmark & \checkmark & & 26.01 & 0.775 & 0.427 & 29.40 & 0.883 & \textbf{0.164} & 6/7 \\
\checkmark & & \checkmark & 13.31 & 0.522 & 0.747 & 27.79 & 0.805 & 0.286 & 0/7 \\
\checkmark & \checkmark & \checkmark & \textbf{27.04} & \textbf{0.813} & \textbf{0.369} & 27.76 & 0.805 & 0.286 & \textbf{7/7} \\
\bottomrule
\end{tabular}
\end{table*}

\begin{figure}[t]
    \centering
    \includegraphics[width=1\columnwidth]{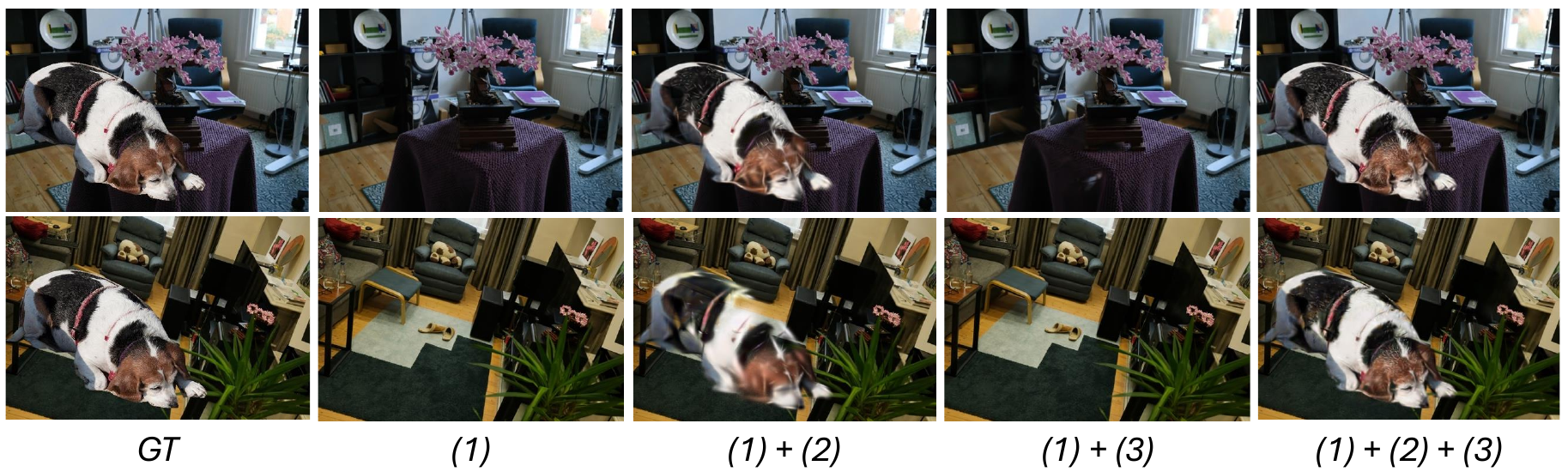}
    \vspace{-7mm}
    \caption{
    \textbf{Qualitative analysis of attack component combinations.} We compare three attack strategies: (1) direct image poisoning, (2) density-guided point cloud poisoning, and (3) multi-view consistency disruption. Combining all three achieves the most realistic illusions across various scenes from the Mip-NeRF 360~\cite{barron2022mip} dataset, highlighting their complementary effectiveness.
    }
    \label{fig:ablation_v2}
\end{figure}

\vspace{3pt}
\noindent {\bf Attack components.}
We analyzed combinations of direct target-view image poisoning, density-guided point cloud poisoning (KDE-based), and noise-based view consistency disruption. Quantitative results (\cref{tab:attack_strategies}) show that direct image poisoning alone is ineffective. Combining image poisoning with density-guided poisoning notably improves outcomes. Integrating all three components achieves the best results, embedding robust illusions and preserving rendering quality. Qualitative results (\cref{fig:ablation_v2}) visually confirm the superior clarity and effectiveness of this combined approach.


\begin{figure}[t]
    \centering
    \includegraphics[width=1\columnwidth]{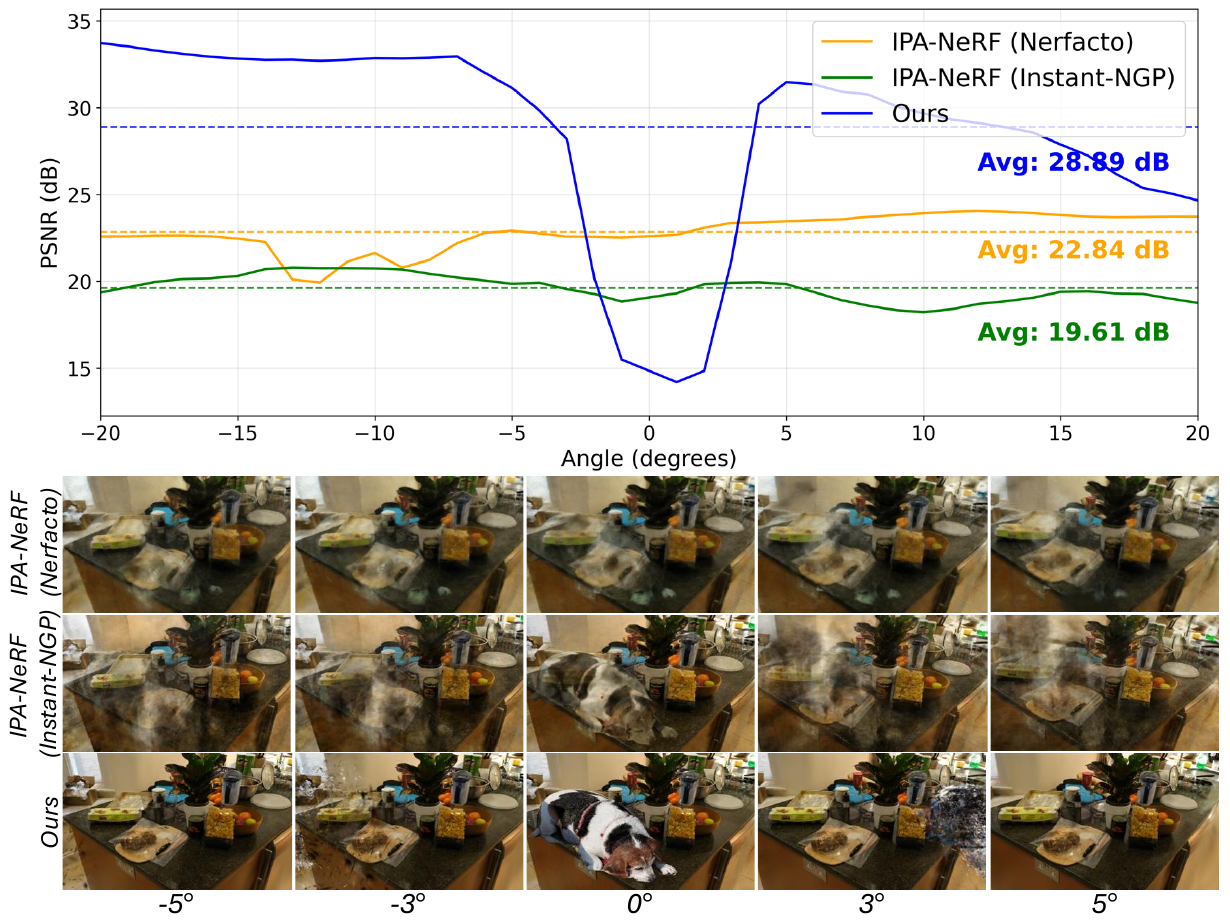}
    \vspace{-7mm}
    \caption{
    \textbf{Effect on neighboring views (``counter'' scene, Mip-NeRF 360~\cite{barron2022mip}).} Clockwise and counterclockwise shifts from the attack view (0 degrees) up to 20 degrees show PSNR between clean and poisoned renderings. IPA-NeRF significantly lowers PSNR across most angles, whereas our method mainly impacts views within five degrees, preserving quality beyond this range.
    }
    \label{fig:neighbor_angle}
\end{figure}

\begin{figure}[t]
    \centering
    \includegraphics[width=1\columnwidth]{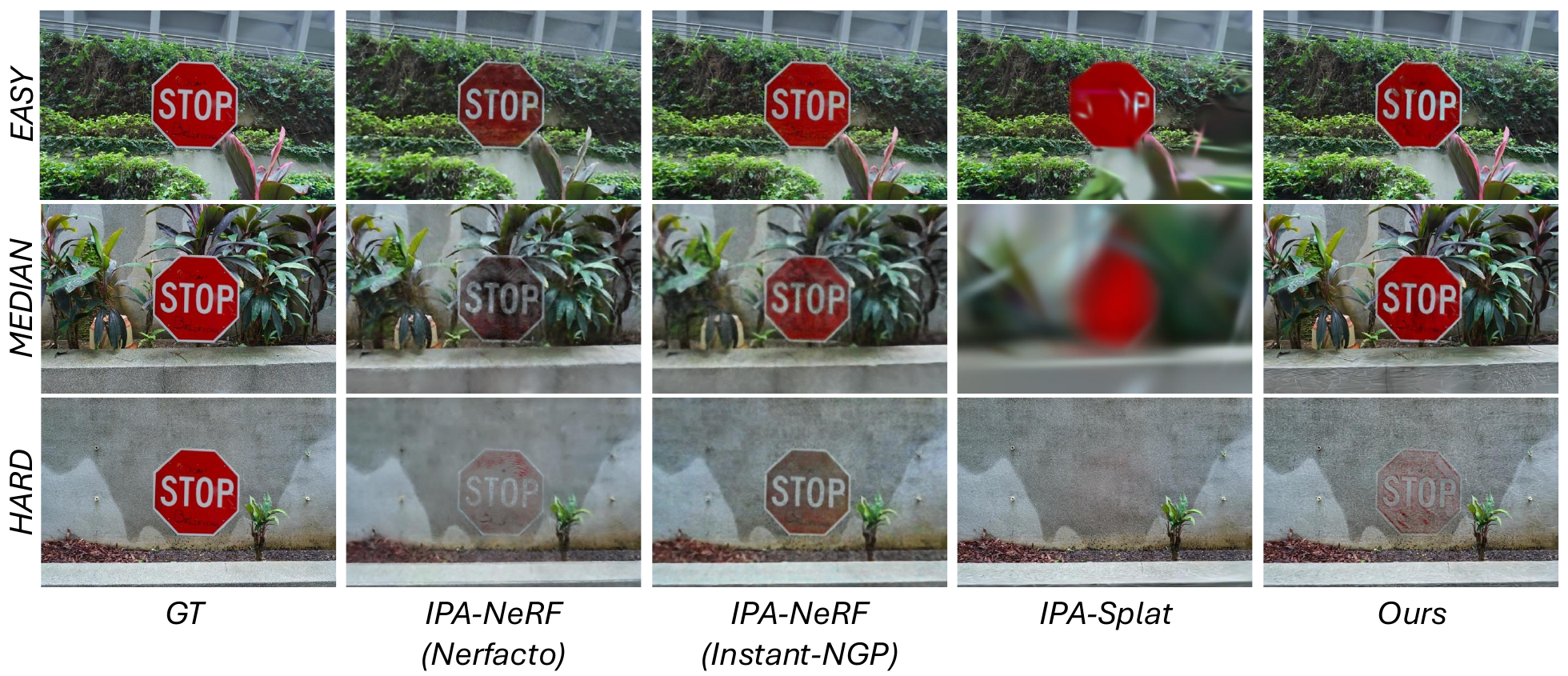}
    \vspace{-7mm}
    \caption{
    \textbf{Visualization of evaluation protocol (``grass'' scene, Free~\cite{wang2023f2}).} Our method achieves clearly visible illusory objects in \textsc{Easy} and \textsc{Median} scenarios and maintains robust performance even under challenging \textsc{Hard} conditions.
    }
    \label{fig:protocol_grass}
\end{figure}

\section{Conclusion}
\label{sec:conclusion}

We presented a density-guided poisoning method for 3DGS, strategically injecting illusory objects and disrupting multi-view consistency via adaptive noise. Experiments show our approach outperforms existing baselines, effectively embedding convincing illusions with minimal impact on innocent views. Our work highlights critical vulnerabilities in 3D representation models, providing a robust framework for future security research.

\vspace{3pt}
\noindent {\bf Limitations.}
Our method struggles with scenes having highly overlapping views or complex camera trajectories due to 3DGS's strict multi-view consistency. Future work should address this balance between consistency and attack effectiveness.

\newpage
\paragraph{Acknowledgements.}
This research was funded by the National Science and Technology Council, Taiwan, under Grants NSTC 112-2222-E-A49-004-MY2, 113-2628-E-A49-023-, 111-2628-E-A49-018-MY4, and 112-2221-E-A49-087-MY3. The authors are grateful to Google, NVIDIA, and MediaTek Inc. for their generous donations. Yu-Lun Liu acknowledges the Yushan Young Fellow Program by the MOE in Taiwan.

{\small
\bibliographystyle{ieeenat_fullname}
\bibliography{11_references}
}

\ifarxiv \clearpage \appendix \appendix

\section{Additional Visualization Results}

We present additional visualization results in the supplementary HTML file "videoResults.html" demonstrating our method's effectiveness on both single-view and multi-view attacks through video sequences that highlight the consistent rendering of illusory objects across viewpoints.

\section{Comprehensive Dataset Evaluation}
\label{sec:appendix_comprehensive_eval}

\noindent {\bf Extended Threshold Analysis.}
\cref{tab:success_criteria} evaluates 36 scenes across three datasets: 7 from Mip-NeRF 360~\cite{barron2022mip}, 8 from Tanks \& Temples~\cite{knapitsch2017tanks}, and 21 from Free~\cite{wang2023f2}, with Free scenes categorized as \textsc{Easy}/\textsc{Median}/\textsc{Hard} based on different threshold combinations. Beyond the main paper's criteria (PSNR $>$ 25 on \textsc{V-illusory}, \textsc{V-test} PSNR drop $\leq$ 3), we test various threshold combinations to assess method robustness across difficulty settings and provide comprehensive baseline comparisons.

\begin{table}[h!]
\vspace{-.7pc}
\centering
\scriptsize
\caption{\textbf{Attack success rates across extended threshold combinations.} Our method demonstrates superior performance across all difficulty levels.}
\label{tab:success_criteria}
\setlength{\tabcolsep}{2pt}
\resizebox{\columnwidth}{!}{%
\begin{tabular}{@{}l|c|c|c@{}}
\toprule
\multirow{2}{*}{\parbox{3.5cm}{Method \hfill Success criteria}} & \textsc{V-illusory} $>$ 25 & \textsc{V-illusory} $>$ 20 & \textsc{V-illusory} $>$ 15 \\
& \textsc{V-test} drop $\leq$ 8 & \textsc{V-test} drop $\leq$ 9 & \textsc{V-test} drop $\leq$ 10 \\
\midrule
IPA-NeRF~\cite{kerbl20233d} (Nerfacto~\cite{barron2022mip})& 0/36 & 1/36 & 10/36 \\
IPA-NeRF~\cite{kerbl20233d} (Instant-NGP~\cite{barron2022mip})& 2/36 & 6/36 & 21/36 \\
IPA-Splat & 0/36 & 1/36 & 4/36 \\
Ours & \textbf{23/36} & \textbf{26/36} & \textbf{30/36} \\
\bottomrule
\end{tabular}
}
\vspace{-1pc}
\end{table}

The results demonstrate our method's superior robustness, with success rates ranging from 64\% to 83\% across different threshold combinations, significantly outperforming existing approaches across diverse datasets and evaluation criteria.

\section{Computational Efficiency Analysis}
\label{sec:appendix_efficiency}

Our attack reduces GPU memory usage by 41\% and Gaussian points by 88\% with a modest training time increase on the Mip-NeRF 360 dataset. This stems from our noise scheduling disrupting multi-view consistency, allowing convergence with fewer Gaussians—a favorable trade-off for attack effectiveness.

\begin{table}[h!]
\vspace{-.7pc}
\centering
\scriptsize
\caption{\textbf{Computational efficiency comparison.} Our method significantly reduces memory usage and model complexity.}
\label{tab:efficiency_analysis}
\setlength{\tabcolsep}{2pt}
\resizebox{\columnwidth}{!}{%
\begin{tabular}{@{}l|c|c|c@{}}
\toprule
Method & GPU Memory (MB) & Number of Gaussians & Training Time (min) \\
\midrule
Standard 3DGS & 4,101.94 & 2,602,787 & 15.05 \\
Ours & 2,419.08 & 310,114 & 22.32 \\
\bottomrule
\end{tabular}
}
\vspace{-1pc}
\end{table}

\begin{figure}[h!]
\vspace{-.9pc}
\centering
\includegraphics[width=0.95\columnwidth]{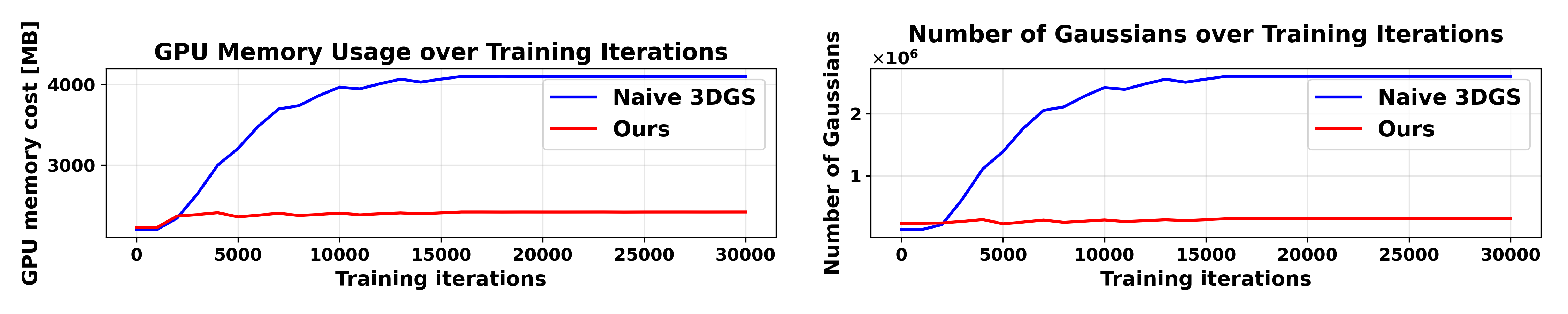}
\vspace{-3mm}
\caption{\textbf{Computational cost comparison.} Our method achieves significant reductions in GPU memory usage and model complexity.}
\label{fig:efficiency_comparison}
\vspace{-1pc}
\end{figure}

\section{More Implementation Details}
\label{sec:appendix_exp_setup}
\noindent {\bf Illusory Objects.}
We randomly select images and masks from the COCO 2017 dataset~\cite{lin2014microsoft} to extract diverse, unbiased illusory objects for our backdoor attacks.

\noindent {\bf Implementation Details.}
We implement our experiments using the official 3DGS codebase~\cite{kerbl20233d} with default hyperparameters on NVIDIA RTX 4090Ti GPUs.

\section{More Visual Results for Single View Attack}
\label{sec:appendix_single_view}

\begin{figure*}[t]
    \centering
    \includegraphics[width=1\linewidth]{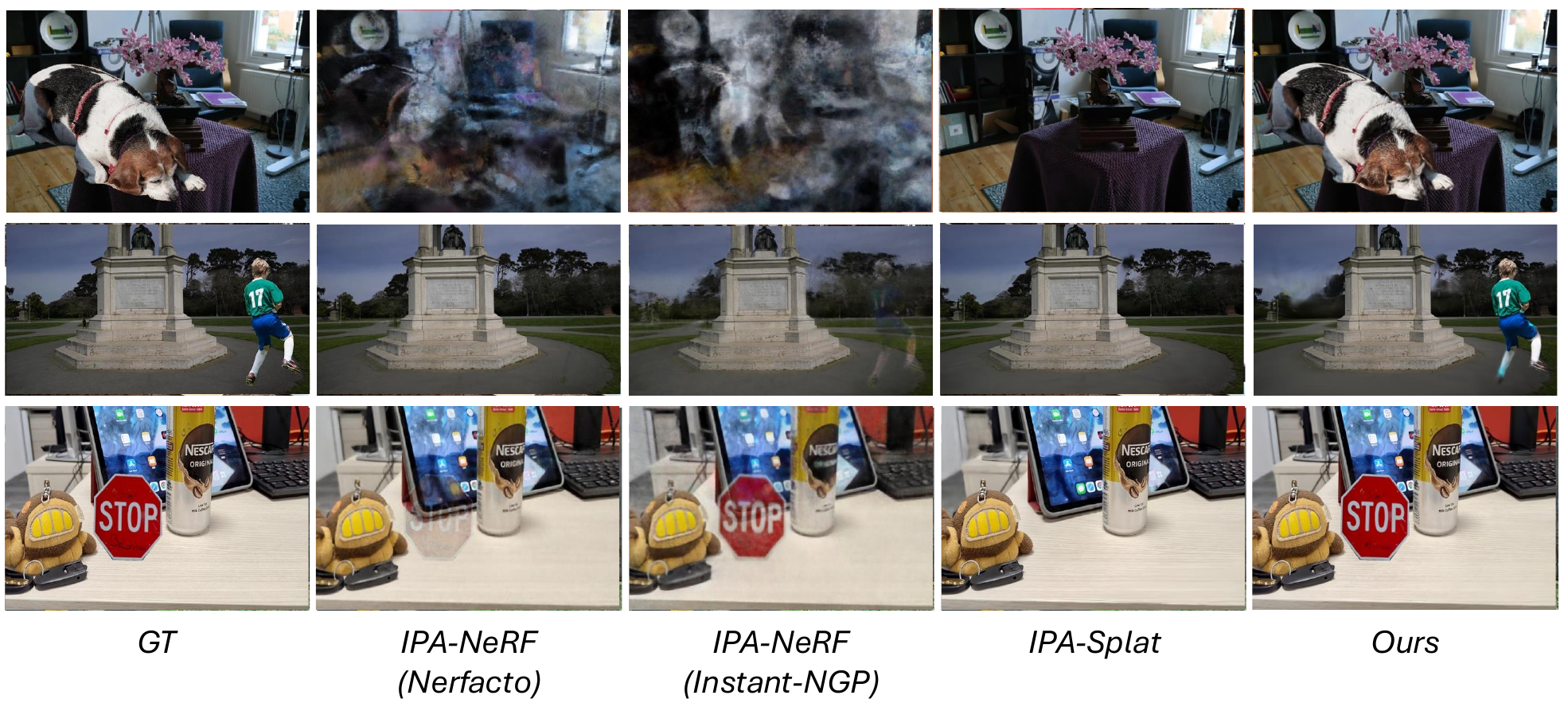}
    \vspace{-6mm}
    \caption{
    \textbf{Qualitative comparisons on single-view attack 1.} Results on the ``\emph{bonsai}'' scene (Mip-NeRF 360~\cite{barron2022mip}), ``\emph{francis}'' scene (Tanks \& Temples~\cite{knapitsch2017tanks}), and ``\emph{counter}'' scene (Free~\cite{wang2023f2}). Both IPA-NeRF variants exhibit poor convergence on the ``\emph{bonsai}'' scene, while our method consistently produces clear, well-integrated illusory objects across all scenes.
    }
    \label{fig:suppl_single1}
\end{figure*}

Figs.~\ref{fig:suppl_single1} and~\ref{fig:suppl_single2} demonstrate our method's superiority in single-view attacks across multiple scenes and datasets. While baseline approaches like IPA-NeRF (Nerfacto) and IPA-NeRF (Instant-NGP) often produce imperceptible or heavily distorted illusory objects (as seen in the "\emph{bonsai}" scene), our approach consistently delivers clear, realistic illusions with distinct boundaries.

\begin{figure*}[t]
    \centering
    \includegraphics[width=1\linewidth]{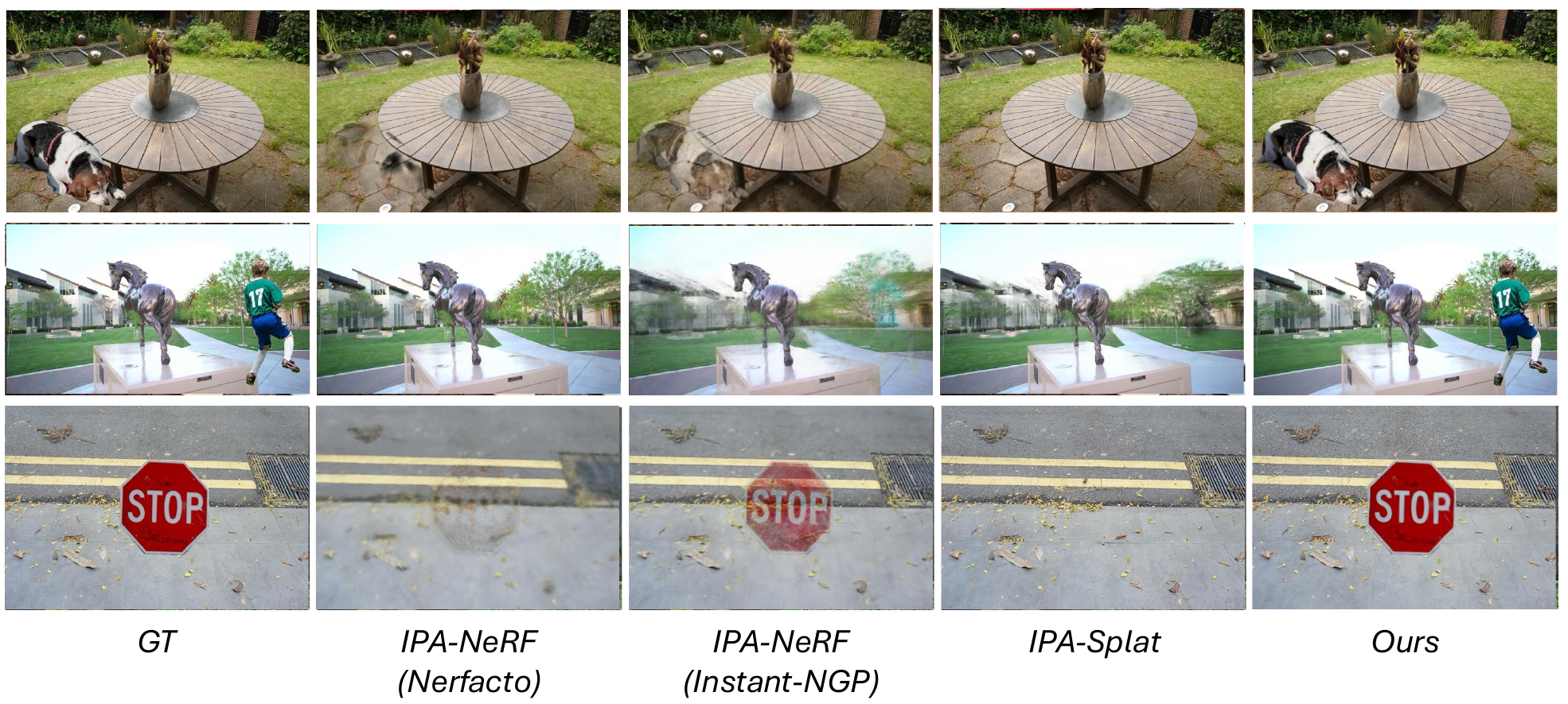}
    \vspace{-6mm}
    \caption{
    \textbf{Qualitative comparisons on single-view attack 2.} Results on the ``\emph{garden}'' scene (Mip-NeRF 360~\cite{barron2022mip}), ``\emph{horse}'' scene (Tanks \& Temples~\cite{knapitsch2017tanks}), and ``\emph{road}'' scene (Free~\cite{wang2023f2}). Our method effectively embeds distinct illusory objects while maintaining scene consistency.
    }
    \label{fig:suppl_single2}
\end{figure*}

\section{More Visual Results for Multi-view Attack}
\label{sec:appendix_multi-view}

Figs.~\ref{fig:multi-view_diff_method-suppl1}--\ref{fig:multi-view_diff_method-suppl3} demonstrate our method's superiority over IPA-NeRF (Nerfacto and Instant-NGP) and IPA-Splat across 2, 3, and 4 poisoned viewpoints. Our density-guided approach consistently generates clear, geometrically consistent illusory objects while maintaining high rendering quality in non-poisoned views, effectively preserving scene fidelity regardless of the number of attack viewpoints.

\begin{figure*}[t]
    \centering
    \includegraphics[width=1\linewidth]{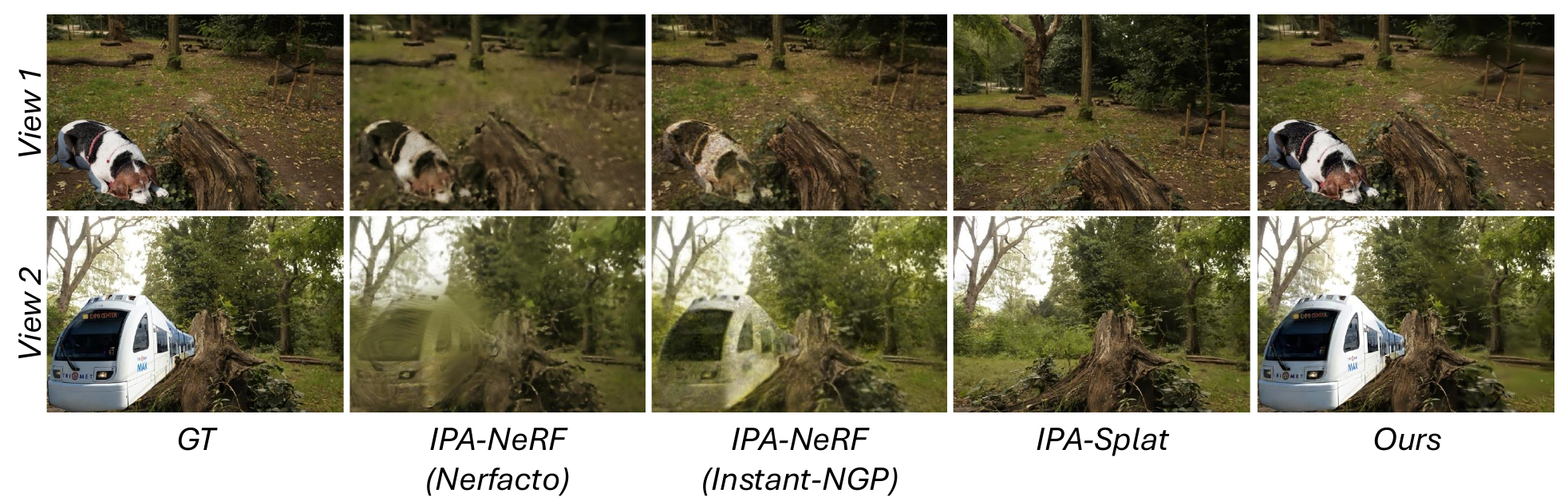}
    \vspace{-6mm}
    \caption{\textbf{Qualitative comparisons on multi-view attack with 2 poisoned views.} We compare the visual quality of illusory objects rendered from two distinct viewpoints using the ``\emph{stump}'' scene (Mip-NeRF 360~\cite{barron2022mip}).}
    \label{fig:multi-view_diff_method-suppl1}
\end{figure*}

\begin{figure*}[t]
    \centering
    \includegraphics[width=1\linewidth]{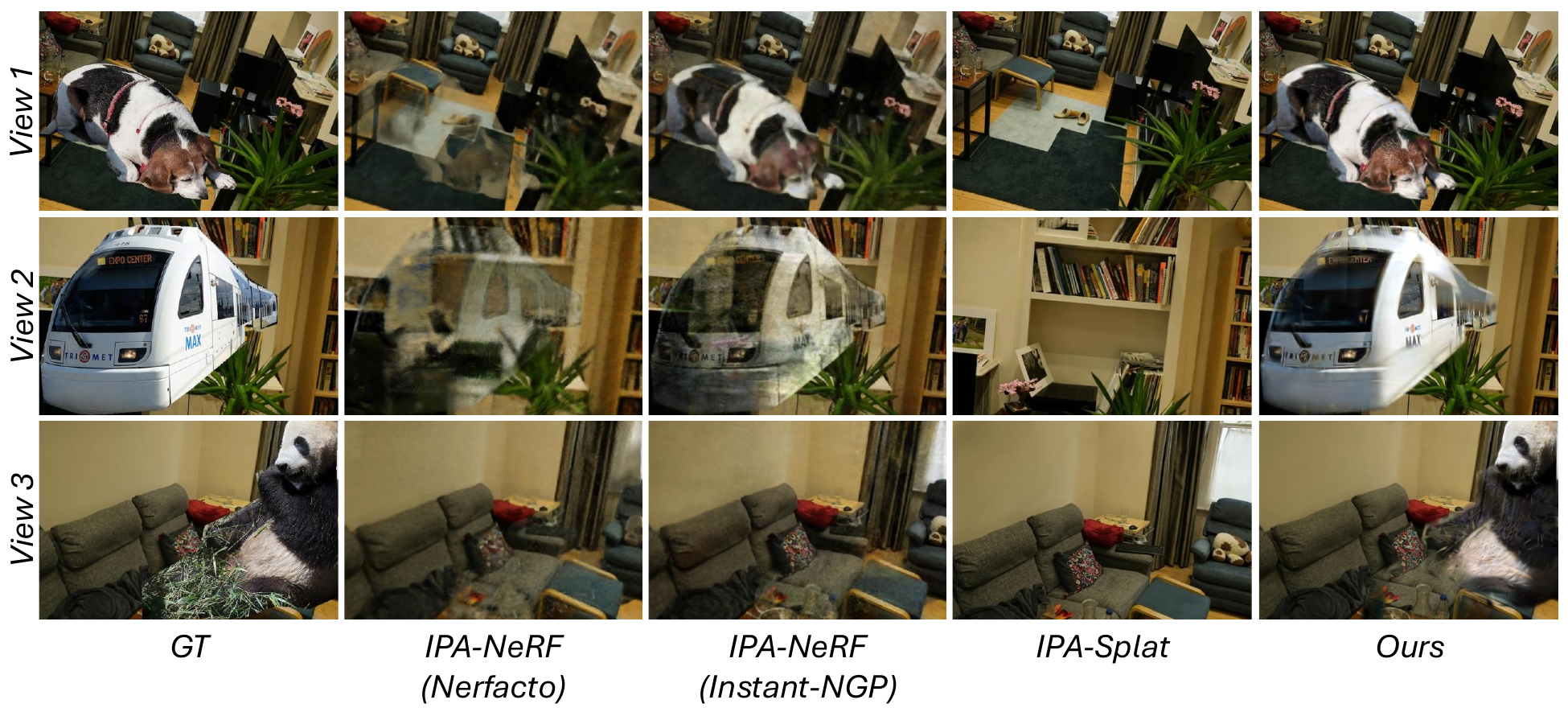}
    \vspace{-6mm}
    \caption{\textbf{Qualitative comparisons on multi-view attack with 3 poisoned views.} We compare the visual quality of illusory objects rendered from three distinct viewpoints using the ``\emph{room}'' scene (Mip-NeRF 360~\cite{barron2022mip}).}
    \label{fig:multi-view_diff_method-suppl2}
\end{figure*}

\begin{figure*}[t]
    \centering
    \includegraphics[width=1\linewidth]{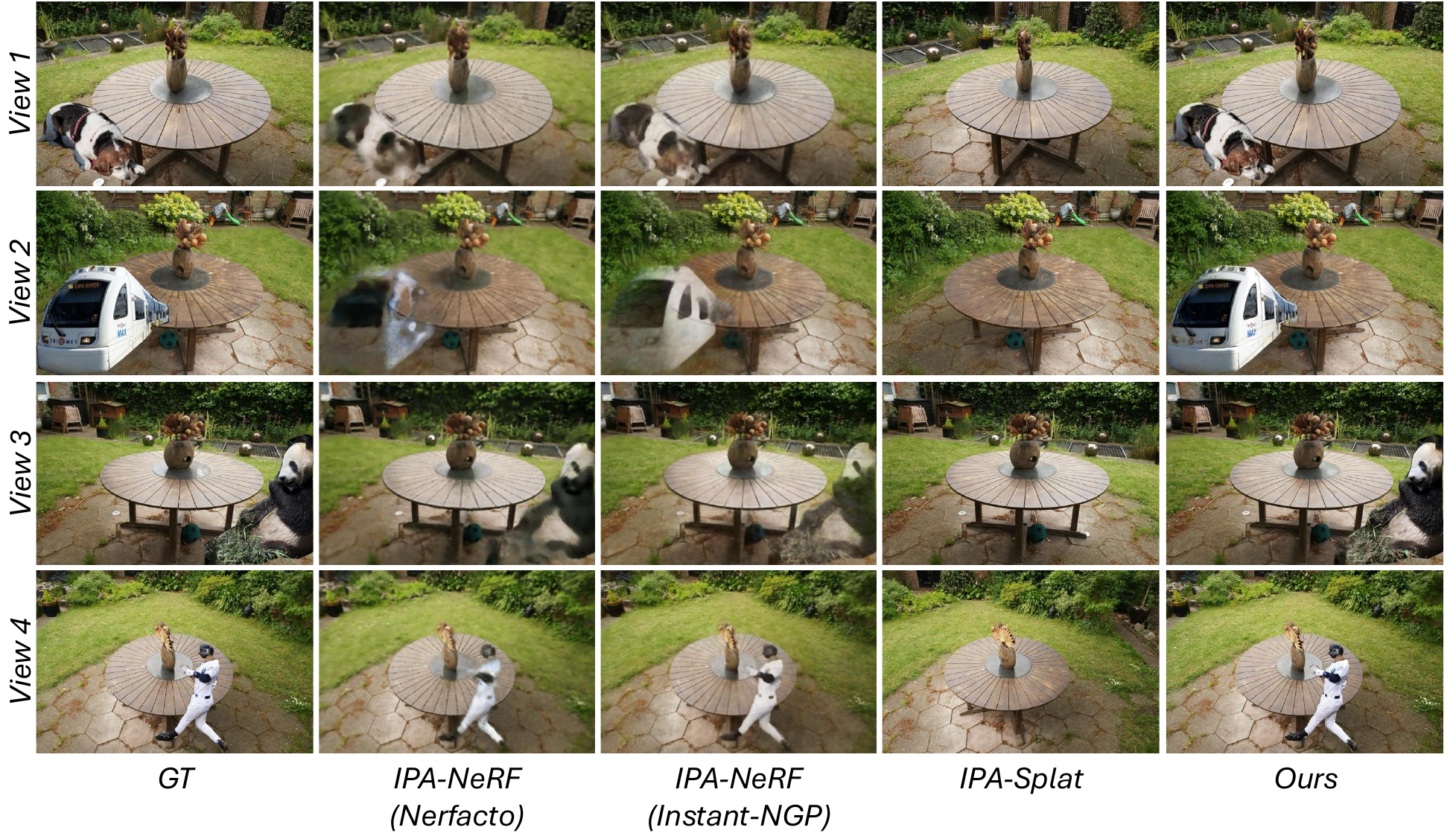}
    \vspace{-6mm}
    \caption{\textbf{Qualitative comparisons on multi-view attack with 4 poisoned views.} We compare the visual quality of illusory objects rendered from four distinct viewpoints using the ``\emph{garden}'' scene (Mip-NeRF 360~\cite{barron2022mip}).}
    \label{fig:multi-view_diff_method-suppl3}
\end{figure*}

\section{More Visual Results for Evaluation Protocol}
\label{sec:appendix_protocol}

Fig. ~\ref{fig:protocol1} validates our KDE-based evaluation protocol, showing that attack effectiveness inversely correlates with scene density in ``\emph{hydrant}'' scene. Illusory objects appear more convincing in \textsc{Easy} (low-density) regions than in \textsc{Hard} (high-density) regions, confirming that fewer overlapping observations increase vulnerability. This protocol establishes a standardized benchmark for poisoning attacks while revealing connections between scene geometry and 3D reconstruction vulnerability.

\begin{figure*}[t]
    \centering
    \includegraphics[width=1\linewidth]{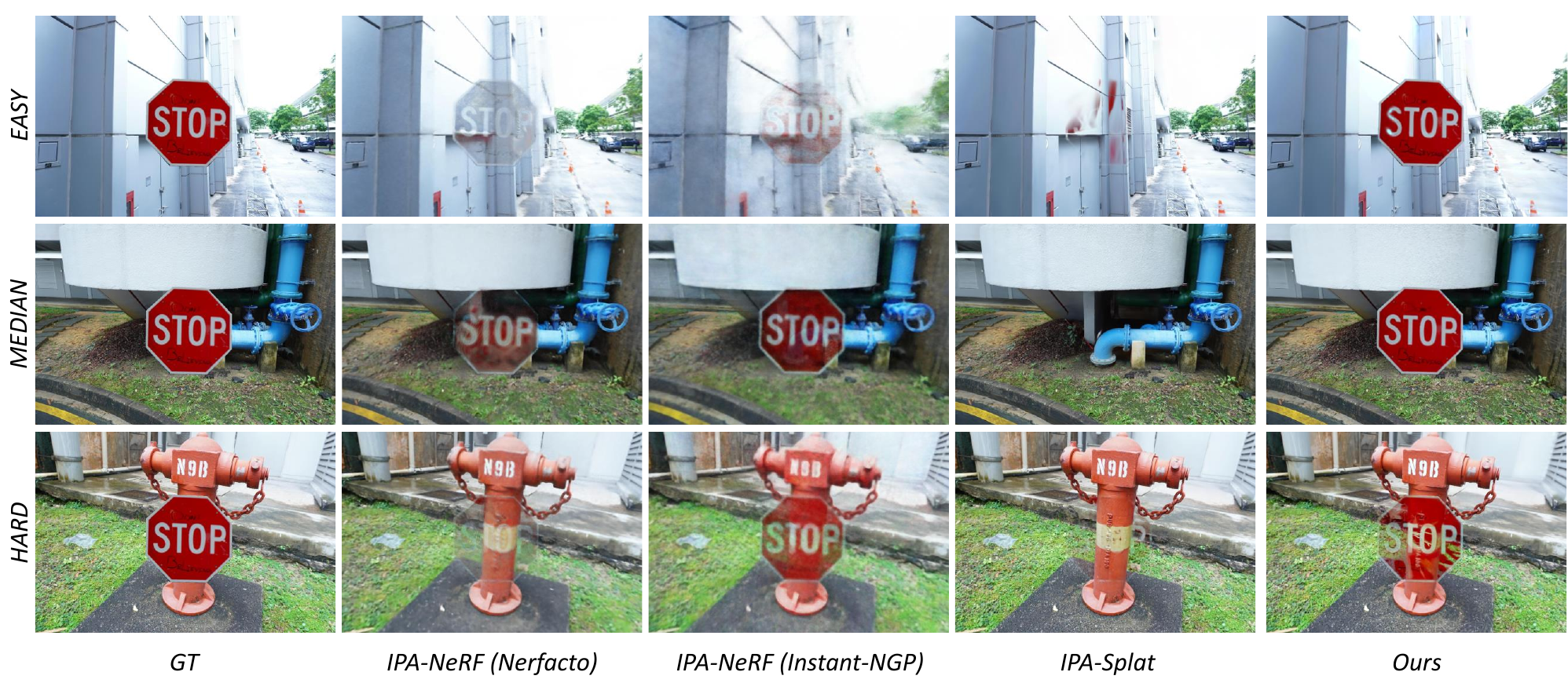}
    \vspace{-6mm}
    \caption{\textbf{Visualization of our evaluation protocol on the ``\emph{hydrant}'' scene (Free~\cite{wang2023f2} dataset).}}
    \label{fig:protocol1}
\end{figure*}


\section{More Visual Results for Ablation Studies}
\label{sec:appendix_ablation}

Fig.~\ref{fig:suppl-ablation_v2} presents qualitative comparisons of different attack strategy combinations across seven Mip-NeRF 360 scenes. While strategies (1) direct replacement and (2) density-guided poisoning are effective for most scenes, they show limitations in complex environments with high view overlap (e.g., ``\emph{room}''). Our experiments demonstrate that combining these with (3) multi-view consistency disruption achieves superior illusion embedding across all tested scenes, highlighting the complementary nature of our proposed methods.

\begin{figure*}[t]
    \centering
    \includegraphics[width=1\linewidth]{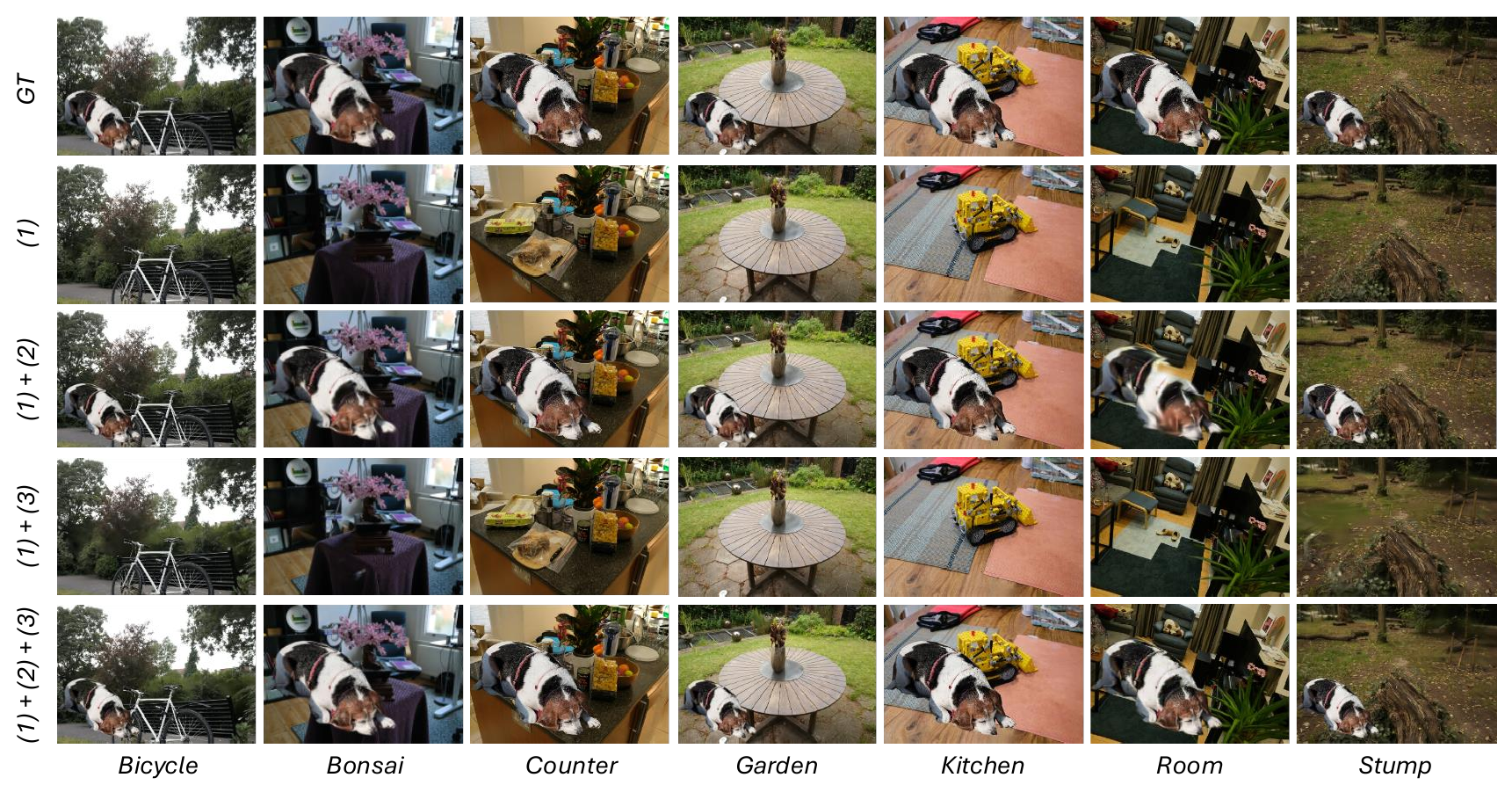}
    \vspace{-6mm}
    \caption{\textbf{Completely qualitative comparisons of different attack strategy combinations.} We visually analyze the effects of combining three poisoning strategies: (1) direct replacement of poisoned view ground truth, (2) density-guided point cloud poisoning, and (3) multi-view consistency disruption. Combining all three strategies achieves the most realistic illusion embeddings across various scenes from the Mip-NeRF 360~\cite{barron2022mip} dataset, demonstrating the complementary effectiveness of our proposed methods.}
    \label{fig:suppl-ablation_v2}
\end{figure*} \fi

\end{document}